\documentclass[lettersize,journal]{IEEEtran}
\usepackage{amsmath,amsfonts}
\usepackage{algorithmic}
\usepackage{algorithm}
\usepackage{array}
\usepackage[caption=false,font=normalsize,labelfont=sf,textfont=sf]{subfig}
\usepackage{textcomp}
\usepackage{stfloats}
\usepackage{url}
\usepackage{verbatim}
\usepackage{graphicx}
\usepackage{cite}

\usepackage{booktabs}
\usepackage{multirow}
\usepackage{bbding}
\usepackage{amsmath}
\usepackage{overpic}
\usepackage{amsfonts}
\usepackage{mathtools}
\usepackage{empheq}
\usepackage[table]{xcolor}

\usepackage{utfsym}

\begin{document}

\title{Source-Free Object Detection with Detection Transformer}

\author{Huizai Yao, Sicheng Zhao, \IEEEmembership{Senior Member, IEEE}, Shuo Lu, Hui Chen, Yangyang Li, Guoping Liu, Tengfei Xing, \\Chenggang Yan, Jianhua Tao, Guiguang Ding, \IEEEmembership{Senior Member, IEEE}


\IEEEcompsocitemizethanks{
\IEEEcompsocthanksitem H. Yao, S. Zhao, H. Chen, J. Tao, and G. Ding are with BNRist, Tsinghua University, Beijing 100084, China. J. Tao is also with the Department of Automation, Tsinghua University. G. Ding is also with the School of Software, Tsinghua University (e-mail: victoryaohz@gmail.com, schzhao@tsinghua.edu.cn, jichenhui2012@gmail.com, jhtao@tsinghua.edu.cn, dinggg@tsinghua.edu.cn). (Corresponding author: Sicheng Zhao)\protect
\IEEEcompsocthanksitem S. Lu is with the Institute of Automation, Chinese Academy of Sciences, China (e-mail: lushuo2024@ia.ac.cn).\protect
\IEEEcompsocthanksitem Y. Li is with the Academy of Cyber, China (e-mail: liyangyang@ict.ac.cn).\protect
\IEEEcompsocthanksitem G. Liu and T. Xing are with DiDi Chuxing, China (e-mail: liuguoping@didiglobal.com, xingtf@foxmail.com).\protect
\IEEEcompsocthanksitem C. Yan is with the School of Automation, Hangzhou Dianzi University, Hangzhou 310018, China, and Lishui Institute of Hangzhou Dianzi University, Lishui, China (e-mail: cgyan@hdu.edu.cn).
}

}

\markboth{IEEE Transactions on Image Processing, 2025}%
{Shell \MakeLowercase{\textit{et al.}}: A Sample Article Using IEEEtran.cls for IEEE Journals}


\maketitle

\begin{abstract}
Source-Free Object Detection (SFOD) enables knowledge transfer from a source domain to an unsupervised target domain for object detection without access to source data. Most existing SFOD approaches are either confined to conventional object detection (OD) models like Faster R-CNN or designed as general solutions without tailored adaptations for novel OD architectures, especially Detection Transformer (DETR). In this paper, we introduce \underline{F}eature \underline{R}eweighting \underline{AN}d \underline{C}ontrastive Learning Networ\underline{K} (FRANCK), a novel SFOD framework specifically designed to perform query-centric feature enhancement for DETRs. FRANCK comprises four key components: (1) an Objectness Score-based Sample Reweighting (OSSR) module that computes attention-based objectness scores on multi-scale encoder feature maps, reweighting the detection loss to emphasize less-recognized regions; (2) a Contrastive Learning with Matching-based Memory Bank (CMMB) module that integrates multi-level features into memory banks, enhancing class-wise contrastive learning; (3) an Uncertainty-weighted Query-fused Feature Distillation (UQFD) module that improves feature distillation through prediction quality reweighting and query feature fusion; and (4) an improved self-training pipeline with a Dynamic Teacher Updating Interval (DTUI) that optimizes pseudo-label quality. By leveraging these components, FRANCK effectively adapts a source-pre-trained DETR model to a target domain with enhanced robustness and generalization. Extensive experiments on several widely used benchmarks demonstrate that our method achieves state-of-the-art performance, highlighting its effectiveness and compatibility with DETR-based SFOD models.
\end{abstract}

\begin{IEEEkeywords}
Transfer Learning, Object Detection, Source-Free Domain Adaptation, Contrastive Learning.
\end{IEEEkeywords}

\begin{figure}[!t]
\begin{center}
\includegraphics[width=\linewidth]{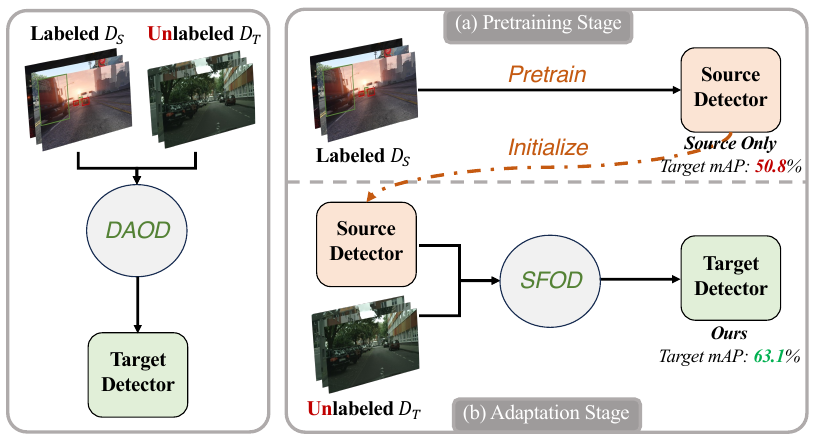}
\caption{Illustration of SFOD setting. \textbf{\textit{Left:}} Conventional Domain Adaptive Object Detection (DAOD) approaches utilize both labeled source Domain ($D_S$) and unlabeled target domain ($D_T$) to transfer the detector to the target domain. \textbf{\textit{Right:}} Source-Free Object Detection (SFOD) adapts source pre-trained model to the target domain when source data is unavailable.}
\label{fig:SFOD}
\end{center}
\end{figure}

\section{Introduction}
\IEEEPARstart{I}{n} the past decade, deep neural networks~\cite{krizhevsky2012imagenet,he2016deep} have significantly advanced object detection (OD). However, state-of-the-art detectors, such as Faster R-CNN~\cite{ren2015faster} and DEtection TRansformer~\cite{carion2020end} (DETR), require large-scale, high-quality labeled data to achieve optimal performance. Collecting and annotating such data is often expensive and labor-intensive. Furthermore, real-world scenarios frequently exhibit domain shift, where the training data or source domain distribution differs from test data or target domain distribution~\cite{amodei2016concrete}. This shift severely degrades the generalization ability of conventional OD models, which has led to extensive research on unsupervised domain adaptive object detection (DAOD)~\cite{rodriguez2019domain,mattolin2023confmix,li2023sigma++,liu2023cigar,hsu2020every,zhou2022multi,wang2021exploring,yu2022mttrans,zhang2023detr,weng2024mean,he2023bidirectional,zeng2024enhancing}. Most DAOD methods rely on adversarial feature alignment\cite{li2023sigma++,hsu2020every} or intermediate domain generation\cite{rodriguez2019domain,mattolin2023confmix}, both of which require access to annotated source data. However, in real-world applications, source data is often unavailable due to privacy concerns or transmission constraints~\cite{liang2024comprehensive}. In such cases, traditional DAOD techniques become infeasible, necessitating source-free object detection (SFOD) as an alternative.

As illustrated in Fig.~\ref{fig:SFOD}, SFOD tackles domain adaptation without access to labeled source data. Instead, it relies solely on a source-pre-trained model for adaptation to the target domain~\cite{li2021free,vs2023instance,li2022source,chu2023adversarial,sinha2023test,deng2024balanced,zhao2024multi,khanh2024dynamic}. Due to the absence of source data, most SFOD methods employ a Mean Teacher\cite{tarvainen2017mean} framework with pseudo-labeling to facilitate adaptation. While these approaches have shown promise, they predominantly focus on Faster R-CNN architectures, leveraging components such as RPNs~\cite{chen2023exploiting, li2022source}. Consequently, they lack critical insights into adapting DETR-based models~\cite{chu2023adversarial,deng2024balanced}. Recent works~\cite{suzuki2023onda, sinha2023test} and a concurrent study~\cite{khanh2024dynamic} have begun exploring SFOD for DETR. However, these efforts either overlook DETR-specific architectural components\cite{suzuki2023onda, sinha2023test} or focus excessively on teacher-student optimization\cite{khanh2024dynamic}, failing to fully exploit DETR’s unique features. Addressing these gaps, we propose a DETR-oriented SFOD framework that effectively incorporates DETR-specific designs for robust adaptation.

To address the source‑free adaptation challenges faced by DETR, we propose \underline{F}eature \underline{R}eweighting \underline{AN}d \underline{C}ontrastive Learning Networ\underline{K} (FRANCK), a unified \textbf{query‑centric} framework that enhances DETR’s adaptation capabilities. We explicitly decompose these challenges into three interconnected levels of alignment: \textbf{category‑level alignment} (mitigating inter‑class confusion), \textbf{instance‑level alignment} (balancing and supervising samples via pseudo‑labels), and \textbf{feature‑level alignment} (stabilizing cross‑domain feature transfer). Guided by this perspective, each module in FRANCK directly targets one level while sharing a common query‑centric interface.
CMMB enhances category‑level alignment by performing class‑wise contrastive learning with matching‑based memory banks to improve query discriminability. OSSR addresses instance‑level alignment by dynamically reweighting query losses through attention‑derived objectness scores, mitigating class imbalance and inadequate supervision. UQFD improves feature‑level alignment by distilling features with uncertainty‑weighted query‑fused masks, leading to more stable teacher–student transfer.

\begin{figure}[!t]
\begin{center}
\includegraphics[width=\linewidth]{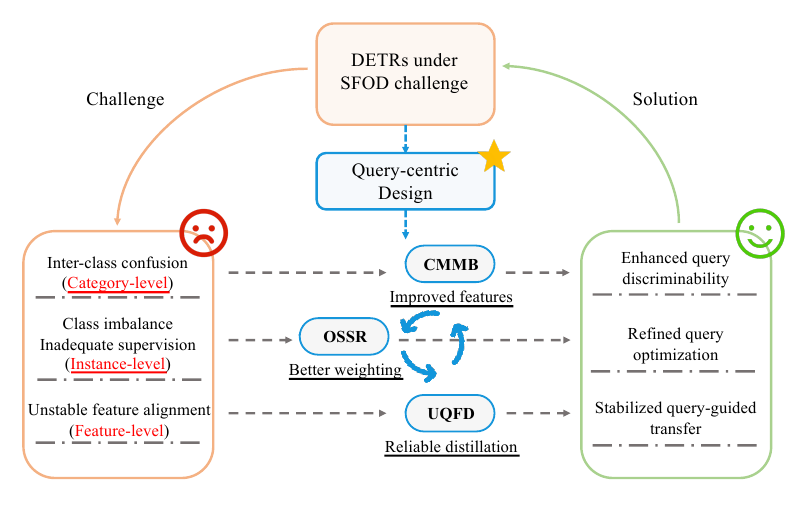}
\caption{A conceptual framework illustrating how FRANCK addresses DETR’s source‑free challenges through a unified query‑centric design. The three challenges are organized as category‑level alignment (inter‑class confusion), instance‑level alignment (class imbalance and inadequate supervision), and feature‑level alignment (unstable feature alignment). Each module (CMMB, OSSR, UQFD) targets one of these challenges, while their shared reliance on query representations forms a synergistic loop where improved features, better weighting, and reliable distillation reinforce each other for robust and efficient adaptation.}
\label{fig:synergy}
\end{center}
\end{figure}

These modules form a coherent pipeline and mutually reinforce one another, as illustrated in Fig.~\ref{fig:synergy}. Stronger query embeddings from CMMB enable more precise sample weighting in OSSR; enriched queries also guide UQFD to produce more reliable distillation masks; and the stable features obtained from UQFD feed back into both CMMB and OSSR. Through this shared reliance on query representations, FRANCK unifies contrastive learning, sample reweighting, and feature distillation across category‑level, instance‑level, and feature‑level alignment, resulting in synergistic improvements in both discriminability and transferability for DETR‑based SFOD.

Our main contributions are as follows:
\begin{itemize}
    \item We systematically explore the challenges of SFOD on DETRs, an area that has received limited attention, and propose a novel framework that explicitly incorporates DETR-specific architectural designs.
    \item We propose FRANCK, a novel framework that introduces several key innovations tailored for DETR-based source-free domain adaptive object detection. Motivated by a query-centric representation enhancement principle, the components cooperate to effectively adapt the model to the target domain and improve detection performance.
    \item We conduct extensive experiments on several widely-used benchmarks, demonstrating that FRANCK achieves state-of-the-art performance in SFOD for DETRs.
\end{itemize}

The remainder of this paper is structured as follows: Section~\ref{sec:related_work} provides a comprehensive review of related work, covering Object Detection (OD), Domain Adaptive Object Detection (DAOD), Source-Free Domain Adaptation (SFDA), and Source-Free Object Detection (SFOD). Section~\ref{sec:proposed_method} details the proposed FRANCK framework and its key components. Section~\ref{sec:experiments} presents experimental results, including quantitative analysis, ablation studies, and visualization experiments, along with essential implementation details. Finally, Section~\ref{sec:conclusion} summarizes our findings and concludes the paper.

\section{Related Work} \label{sec:related_work}
This section provides an overview of research relevant to our method, covering OD, DAOD, SFDA, and SFOD.

\subsection{Object Detection}

Object detection (OD) seeks to identify and localize objects in images. With the rise of deep learning and convolutional neural networks (CNNs) \cite{he2016deep}, OD has made significant progress. Traditional methods are typically categorized into two-stage detectors (e.g., R-CNN \cite{girshick2015fast}, Faster R-CNN \cite{ren2015faster}) that generate region proposals before classification, and one-stage detectors (e.g., SSD \cite{liu2016ssd}, FCOS \cite{tian2019fcos}, YOLO \cite{redmon2016you}) that predict objects directly. While efficient, these models often rely on heuristic components like Non-Maximum Suppression (NMS), making them sensitive to hyperparameters.

Transformer-based models have redefined OD by formulating it as a set prediction problem. DETR (DEtection TRansformer) and its variants \cite{carion2020end, jia2023detrs, zhao2024detrs, liu2022dab, li2022dn, zhu2020deformable, zhang2022dino} remove the need for NMS, enabling end-to-end detection. Variants such as Deformable DETR \cite{zhu2020deformable} and DN-DETR \cite{li2022dn} improve convergence and robustness. However, DETR-based models often underperform under domain shifts and have limited cross-domain generalization \cite{wang2021exploring, yu2022mttrans, zhang2023detr, weng2024mean}. In particular, DETR-based source-free object detection remains underexplored. To bridge this gap, we adopt DETR as our base and investigate its extension to source-free cross-domain object detection.

\subsection{Domain Adaptive Object Detection}
Domain Adaptive Object Detection (DAOD) aims to mitigate domain shifts and enhance the generalization ability of object detectors. Most DAOD research has focused on traditional detection frameworks such as Faster R-CNN~\cite{chen2018domain, deng2023harmonious, zhang2021rpn, he2020domain, yao2021multi, cao2023contrastive, vs2021mega, li2023igg,li2024react}, FCOS~\cite{hsu2020every, zhou2022multi, li2023sigma++, liu2023cigar}, and YOLO~\cite{zhang2021domain, hnewa2023integrated}. Recently, DAOD methods tailored for DETRs have emerged. For example, SFA~\cite{wang2021exploring} aligns features at both encoder and decoder levels, while MTTrans~\cite{yu2022mttrans} adopts multi-level feature alignment within a Mean Teacher~\cite{tarvainen2017mean} framework. Other approaches, such as DA-DETR~\cite{zhang2023detr}, integrate CTBlender with Split-Merge Fusion and Scale Aggregation Fusion for better alignment. MTM~\cite{weng2024mean} utilizes mask-integrated adversarial alignment and mixed queries to ensure consistent learning, while BiADT~\cite{he2023bidirectional} introduces bi-directional domain alignment with token-wise domain embeddings. Similarly, ACCT~\cite{zeng2024enhancing} employs adversarial alignment, confidence thresholding, and contrastive learning to tackle domain shifts. Despite their effectiveness, these methods rely on source data and labels, making them inapplicable in scenarios where source data is unavailable. Our work addresses this limitation by enabling DETR-based DAOD in a source-free setting.

\subsection{Source-Free Domain Adaptation} 
Conventional Unsupervised Domain Adaptation (UDA) relies on source domain data and labels, limiting its applicability in privacy-sensitive scenarios~\cite{jing2024visually, liang2024comprehensive} and under data transmission constraints~\cite{fang2024source, zhao2024multi}. To address this, Source-Free Domain Adaptation (SFDA) has been introduced, enabling adaptation using only unsupervised target data and a pre-trained source model, without direct access to source data. SFDA has been successfully applied to tasks like image classification~\cite{liang2020we, zhang2023class}, semantic segmentation~\cite{liu2021source}, human pose estimation~\cite{peng2023source}, gesture recognition~\cite{Guo2024SpGestureSD}, and panoramic segmentation~\cite{zheng2024360sfuda++}. It leverages various model fine-tuning strategies, with semi-supervised knowledge distillation in a teacher-student framework~\cite{fang2024source} being a widely adopted approach. Additionally, contrastive learning has proven effective in improving generalization by learning discriminative feature representations~\cite{xia2021adaptive, agarwal2022unsupervised}.

Despite SFDA’s success in classification and segmentation, its direct application to object detection is challenging. Unlike these tasks, object detection requires both classification and localization of multiple objects, demanding specialized architectures like Faster R-CNN~\cite{ren2015faster} and DETR~\cite{carion2020end}. To tackle this, Source-Free Object Detection (SFOD) has emerged, enabling domain adaptation for detection tasks. Our work extends SFOD to DETR-based detectors, an area largely unexplored, enhancing DETR’s adaptability under domain shifts.

\subsection{Source-Free Object Detection}
To tackle the challenge of source-free adaptation in object detection, researchers have developed Source-Free Object Detection (SFOD) methods. SED~\cite{li2021free} improves detection performance using self-entropy descent and mosaic augmentation~\cite{bochkovskiy2020yolov4}, while LODS~\cite{li2022source} employs style transfer modules and multi-level feature alignment to minimize domain discrepancy. IRG-SFDA~\cite{vs2023instance} constructs information relation graphs to enhance knowledge distillation and contrastive learning. AASFOD~\cite{chu2023adversarial} applies adversarial alignment on target self-divided data obtained via Monte-Carlo sampling. Meanwhile, Balanced Teacher (BT)\cite{deng2023harmonious} introduces class-balanced instance selection and progressive target variance minimization to mitigate imbalance issues. DACA\cite{zhao2024multi} extends SFOD to multi-source scenarios by incorporating region proposal fusion, pseudo-label ensembling, and class-wise contrastive learning. However, most of these methods either depend on Faster R-CNN-specific components, such as region proposal networks (RPNs), or fail to consider modern DETR architectures.

Notably, a few studies have explored the feasibility of SFOD on DETRs. TeST~\cite{sinha2023test} develops a two-stage self-training process that adapts the teacher and student networks separately, but lacks specific designs tailored to DETR components that could further enhance performance. A concurrent study, DRU~\cite{khanh2024dynamic}, employs masked image consistency~\cite{hoyer2023mic} and dynamic retraining-updating~\cite{zhao2023towards} for effective SFOD on DETRs. Nevertheless, DRU~\cite{khanh2024dynamic} focuses more on general self-training and updating mechanisms, overlooking effective DETR feature adaptation that could further improve SFOD performance. In contrast, our proposed method is carefully designed for the DETR architecture, enabling efficient feature learning and improving detection performance on DETRs.

\section{Proposed Method} \label{sec:proposed_method}

\subsection{Preliminaries} \label{sec:preliminaries}

\noindent \textbf{Problem setup.} We first introduce the problem setup for SFOD tasks. Unless otherwise specified, we consider SFOD tasks under an unsupervised domain adaptation (UDA) setting, where the target labels are entirely unavailable. In the SFOD setting, there exists a source domain $D_S$ sampled from source distribution $p_S (x_S, y_S)$ and a target domain $D_T$ sampled from target distribution $p_T (x_T, y_T)$, where $x$ denotes image and $y$ denotes corresponding label. We follow the closed-set DA setting, in which $D_S$ and $D_T$ both have $k$ foreground categories. Since source data and distribution are unavailable for adaptation, only a source pre-trained model $\theta_S$ and unsupervised target dataset $X_T = \{x_T^i\}_{i=1}^{N_T}$ are available. Our goal is to perform effective adaptation, obtaining a detection model $\theta_T: x_T \to y_p$ that works well on the target domain. We utilize and focus on DETR~\cite{carion2020end,zhu2020deformable} structure for detection. We use Deformable DETR~\cite{zhu2020deformable} as the base detector and denote the total number of object queries in one DETR model as $n_\mathrm{q}$.

\noindent \textbf{Mean-Teacher-based SFOD.} In many SFDA and SFOD applications, the Mean Teacher (MT) framework~\cite{tarvainen2017mean} is a key self-training approach that enables model adaptation without target-domain supervision~\cite{li2022source,vs2023instance,chu2023adversarial,yu2022mttrans,khanh2024dynamic,zhao2024multi}. Originally designed for semi-supervised learning, MT leverages strong-weak augmentation and consistency regularization for knowledge transfer. A key feature is its Exponential Moving Average (EMA) update, ensuring stable parameter update.

\begin{table}[!t]
  \centering
  \caption{Hyperparameters Setting.}
    \resizebox{\linewidth}{!}{%
    \begin{tabular}{c|c|c}
    \toprule[1.5pt]
    Hyperparam & Explanation & Value \\
    \midrule
    $l_{\mathcal{M}}$ & Maximum length of a memory bank & 100 \\
    $H_a, W_a$ & Height and width of RoIAlign output & 7, 7 \\
    $\alpha_t, \gamma$ & Balancing parameter of Focal Loss & 0.25, 2 \\
    $\beta$ & Smoothing parameter for $\mathcal{L}_{\mathrm{wcls}}$ & 0.2 \\
    $\beta^{'}$ & Smoothing parameter for $\mathcal{W}_\mathrm{q}$ in $\mathcal{L}_{\mathrm{fdis}}$ & 1 \\
    $\tau$ & Smoothing parameter for $\mathcal{L}_{\mathrm{cont}}$ & 0.07 \\
    $\alpha_{\mathrm{EMA}}$ & EMA updating rate & 0.999 \\
    $\omega_1$ & Weighting factor of $\mathcal{L}_{\mathrm{cont}}$ & 0.4 \\
    $\omega_2$ & Weighting factor of $\mathcal{L}_{\mathrm{fdis}}$ & 0.1 \\
    $\delta$ & DTUI base interval & 5 \\ 
    $\delta$ & DTUI base interval (Cross-Scene Adaptation) & 60 \\
    $\epsilon$ & DTUI interval increasing rate & 5 \\
    $c_{\mathrm{thresh}}$ & Confidence threshold for pseudo-labeling & 0.3 \\
    \bottomrule[1.5pt]
    \end{tabular}%
   }
  \label{tab:hyperparam}%
\end{table}%

In SFOD, both the teacher and student models are initialized with an identical source-pre-trained network. During training, a target-domain sample from $D_T$ undergoes strong and weak augmentations, which are then fed into the student and teacher networks, respectively. Without supervision, the teacher generates pseudo-labels by confidence-thresholding weakly augmented sample predictions. The student model then updates its parameters by minimizing the following loss:
\begin{equation}
    \mathcal{L}_{\mathrm{det}} = \mathcal{L}_{\mathrm{cls}} + \mathcal{L}_{\mathrm{reg}} + \mathcal{L}_{\mathrm{aux}},
\label{eq:detr_loss}
\end{equation}
where $\mathcal{L}_{\mathrm{det}}$ is the detection loss of DETRs, $\mathcal{L}_{\mathrm{cls}}$ and $\mathcal{L}_{\mathrm{reg}}$ the classification and regression loss, respectively. $\mathcal{L}_{\mathrm{aux}}$ is the auxiliary loss, if applicable. Following Deformable DETR~\cite{zhu2020deformable}, we adopt Focal Loss~\cite{lin2017focal} as the classification loss. The student parameters $\Theta_{\mathrm{stu}}$ are updated via backpropagation, while the teacher parameters $\Theta_{\mathrm{tea}}$ follow EMA updates:
\begin{subequations}
\begin{empheq}[left=\empheqlbrace]{align}
\Theta_{\mathrm{stu}} &\leftarrow \Theta_{\mathrm{stu}} + \eta\frac{\partial(\mathcal{L}_{\mathrm{stu}})}{\partial\Theta_{\mathrm{stu}}}, 
\label{eq:student_update} \\
\Theta_{\mathrm{tea}} &\leftarrow \alpha_{\mathrm{EMA}}\Theta_{\mathrm{tea}} + (1-\alpha_{\mathrm{EMA}})\Theta_{\mathrm{stu}},
\label{eq:teacher_update}
\end{empheq}
\end{subequations}
where $\eta$ is the student learning rate and $\alpha_{\mathrm{EMA}}$ the EMA update rate. MT provides robust model optimization and adaptation in semi-supervised and cross-domain settings, making it a common baseline for many DAOD and SFOD approaches.

\subsection{Overview}
In this section, we outline our problem setup, the Mean‑Teacher‑based SFOD architecture with its update mechanism, and the key components of FRANCK. CMMB leverages pseudo‑label‑induced bipartite matching to build class‑wise memory banks for contrastive learning, enhancing feature discriminability. OSSR mitigates class imbalance by assigning dynamic instance‑wise loss weights via a query‑fused objectness score. UQFD improves knowledge transfer through uncertainty‑weighted, objectness‑guided feature distillation. Finally, we present the overall training loss and DTUI, which strengthens Mean Teacher robustness by dynamically adjusting the EMA update interval.

\subsection{Contrastive Learning with Matching-based Memory Bank}
While the original MT framework with pseudo-labeling lays a solid foundation for SFOD performance, feature representation remains suboptimal~\cite{vs2023instance,zhao2024multi}. To address this, we adopt class-wise contrastive learning following prior studies~\cite{khosla2020supervised,cao2023contrastive,zhao2024multi} and fuse multi-level decoder query features to enhance learning. Additionally, given the class imbalance problem in OD tasks and inspired by bipartite matching in DETRs, we introduce memory banks and a pseudo bipartite matching strategy for class-wise contrastive learning.

\noindent \textbf{Class-wise Contrastive Learning}. Contrastive learning enhances model discriminability by pulling positive samples closer and pushing negatives apart. We build our contrastive loss on the Supervised Contrastive Loss (SCL)~\cite{khosla2020supervised}. Suppose we have $c$ sample groups, denoted as $\mathcal{K}_{a} = \mathcal{K}_{0} \cup \mathcal{K}_{1} \cup \cdots \cup \mathcal{K}_{c-1}$ where each group corresponds to a category. Following SCL, the class-wise contrastive loss is formulated as:
\begin{equation}
\mathcal{L}_{\mathrm{cont}} = \frac{1}{c}\sum_{i=0}^{c-1}\frac{-1}{|\mathcal{K}_{i}|}\sum_{\mathcal{Q} \in \mathcal{K}_{i}}\sum_{K^+ \in \mathcal{K}_{i}}  \log\frac{\exp (\mathcal{Q} \cdot K^+/\tau)}{\sum_{K \in \mathcal{K}_{a}}\exp (\mathcal{Q} \cdot K/\tau)},
\label{equ:losscontvanilla}
\end{equation}
where $\mathcal{Q}$ is the \textit{contrastive learning query feature} that attracts positive keys and repels negative keys in contrastive learning.

\noindent \textbf{Query Feature Fusion and Memory Bank}. For contrastive learning in object detection, it's intuitive to use class-wise instance-level features directly as contrastive samples. However, this can be ineffective or even detrimental because (1) unlike Faster R-CNN, which generates multiple contrastive samples per object via the anchor mechanism~\cite{ren2015faster,vs2023instance}, DETR aims to assign only one query per object at a time~\cite{carion2020end}; and (2)  real-world class distributions are often imbalanced, leading to significant biases. To mitigate these issues, we adopt a simple yet effective memory bank technique~\cite{wu2018unsupervised,tian2020contrastive,zhao2024multi}. We construct $k+1$ memory banks $\{\mathcal{M}_i\}_{i=0}^{k+1}$, one for each category, including a background memory bank $\mathcal{M}_0$. Background features are considered because DETR assigns queries to different objects and naturally generates diverse negative samples, making background features valuable for contrastive learning. Each memory bank maintains a fixed maximum size $l_{\mathcal{M}}$ and is updated using a First-In-First-Out (FIFO) strategy. The non-empty memory banks are denoted as $\{\mathcal{M}_i\}_{i=0}^{v+1}$, where $\mathcal{M}_0$ corresponds to the background memory bank and $v \leq k$. The contrastive loss with memory banks is:
\begin{equation}
\begin{aligned}
\mathcal{L}_{\mathrm{cont}} = \frac{1}{v}\sum_{i=1}^{v+1}\frac{-1}{|\mathcal{M}_{i}|}
\sum_{\mathcal{Q} \in \mathcal{M}_{i}}\sum_{K^+ \in \mathcal{M}_{i}}\log\frac{\exp (\mathcal{Q} \cdot K^+/\tau)}{\sum_{K \in \mathcal{M}_{a}}\exp (\mathcal{Q} \cdot K/\tau)},
\end{aligned}
\label{equ:losscont}
\end{equation}
where $\mathcal{M}_{a}$ represents all samples across memory banks. In Eq.~(~\ref{equ:losscont}), the class index $i$ starts from 1, as background features are only considered negative samples. To further exploit the semantic information contained in multi-scale transformer features, we fuse multi-scale decoder output features of object queries via scale-wise summation. This strategy enhances the representation of instance-level features, providing richer contextual information for improved contrastive learning.

\begin{figure*}[!t]
\centering
\includegraphics[width=0.9\linewidth]{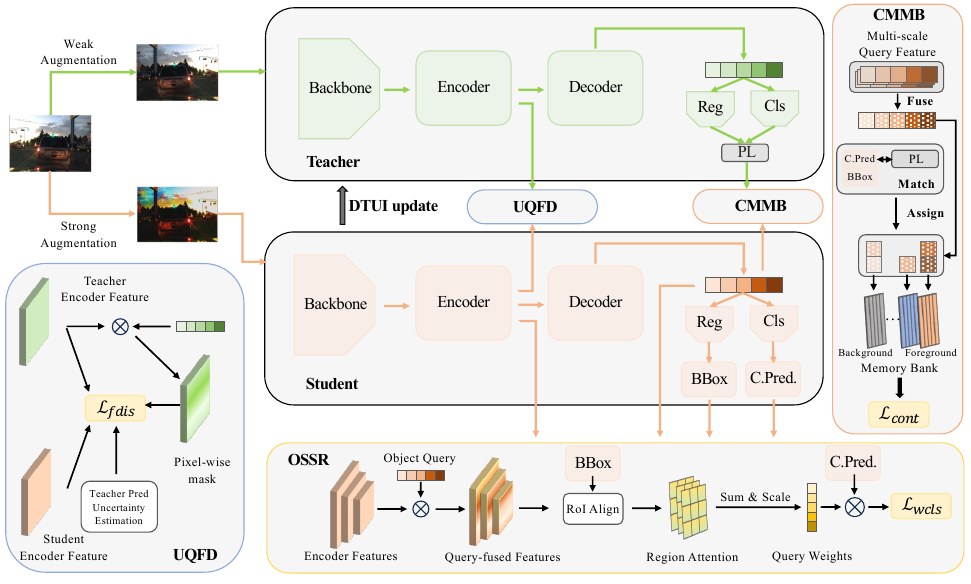}
\caption{The proposed \underline{F}eature \underline{R}eweighting \underline{AN}d \underline{C}ontrastive Learning Networ\underline{K} (FRANCK). Source data is only available at the source pretraining stage. Within FRANCK, the teacher and student models collaborate to optimize the student network through UQFD, OSSR, and CMMB. The teacher network is updated dynamically using DTUI, ensuring more stable and effective adaptation. ``Reg'' and ``Cls'' refer to the regression and classification heads of DETR, respectively. ``BBox'' and ``C.Pred.'' denote the bounding box predictions and classification predictions, respectively, while ``PL'' denotes pseudo-labels.} 
\label{fig:FRANCK}
\end{figure*}

\noindent \textbf{Pseudo Bipartite Matching Assignment}. In the context of contrastive learning for SFOD, a significant challenge stems from the lack of ground truth objects in the target domain, which complicates the construction of contrastive pairs. Due to the set-prediction mechanism of DETRs, RoI-based region features extracted from the backbone~\cite{cao2023contrastive,vs2023instance,zhao2024multi} (\textit{e.g.}, using RoIAlign~\cite{he2017mask}) are decoupled from the instance representations formed by the transformer and object queries. Thus, contrastive learning applied only at the backbone level cannot directly enhance the discriminative power of the object queries, making contrastive learning on object queries a more suitable choice for DETR-based methods. To perform contrastive learning on object queries, a natural approach is to select foreground features from all $n_\mathrm{q}$ query features under DETRs. A straightforward yet suboptimal method is to filter queries based on prediction confidence. Specifically, if a query’s highest probability corresponds to a foreground class and surpasses a predefined threshold, it is considered a foreground feature; otherwise, it is classified as background. However, this approach is heuristic and lacks robustness.

Instead, we leverage \textbf{\textit{bipartite matching}}, one of the core assignment mechanisms in DETRs~\cite{carion2020end}. Since DETRs assign queries to ground-truth labels via bipartite matching, we extend this principle to contrastive pair construction. More specifically, queries matched with pseudo-labels corresponding to an $i$-th foreground class are directly taken as foreground features, while unmatched queries are classified as background. This assignment strategy naturally integrates with DETRs and provides a more stable and unbiased feature selection mechanism. To this end, we adopt bipartite-matching-based contrastive pair assignment between student query features and pseudo-labels as our final strategy for constructing contrastive pairs, as illustrated in Fig.~\ref{fig:FRANCK}. By leveraging pseudo bipartite matching along with the memory bank, we achieve robust class-wise contrastive learning in SFOD for DETRs.

\subsection{Objectness Score Sample Reweighting} \label{ssec:OSSR}

Class imbalance is a common challenge in object detection tasks. A common solution is soft sampling, \textit{i.e.}, reweighting based on prediction quality~\cite{lin2017focal,li2020generalized} or IoU quality~\cite{wang2022soft}. However, in SFOD tasks, the absence of ground-truth labels makes it difficult to estimate prediction quality accurately.

To overcome this, inspired by~\cite{gupta2022ow,chang2023detrdistill,mullappilly2024semi}, we adopt the \textbf{\textit{objectness score}}, which leverages intrinsic feature properties of feature maps. Following these approaches, we first extract and normalize multi-scale encoder features $\{\mathcal{F}_\mathrm{e}^i | i = 3,4,5\}$ and decoder query features $\mathcal{F}_\mathrm{q}$, where $\mathcal{F}_\mathrm{e}^i \in \mathbb{R}^{H_i\times W_i\times C}$ and $\mathcal{F}_\mathrm{q} \in \mathbb{R}^{n_\mathrm{q} \times C}$, together with query bounding boxes as $bbox \in \mathbb{R}^{n_\mathrm{q} \times 4}$. To focus encoder features on query-relevant object information, we perform query-wise fusion and summation to obtain query-fused encoder features $\{\mathcal{F}_{\mathrm{eq}}^i | i = 3,4,5\}$:
\begin{equation}
\mathcal{F}_{\mathrm{eq}}^i = \frac{1}{n_\mathrm{q}} \sum_{j=0}^{n_\mathrm{q}-1} \mathcal{F}_\mathrm{e}^i \cdot \mathcal{F}_\mathrm{q}^T[j] , 
\label{equ:feq}
\end{equation}
where $\mathcal{F}_{\mathrm{eq}}^i \in \mathbb{R}^{ H_i\times W_i}$represents the fused encoder feature across all $n_\mathrm{q}$ queries. Note that since the teacher model in the Mean Teacher (MT) framework undergoes a more stable updating process and encodes more robust features compared student, we leverage $\mathcal{F}_\mathrm{e}^i$ from the teacher encoder to ensure stable and reliable knowledge transfer. Once the attention map is obtained, prior works~\cite{gupta2022ow,mullappilly2024semi} typically upsample the feature map to the original image scale and pooled bounding box features using approximate integer coordinates. However, this naïve approach can lead to information loss and inadequate extraction of small object features, introducing significant bias. Instead, we adopt RoIAlign~\cite{he2017mask}, which uses bilinear interpolation~\cite{jaderberg2015spatial} for precise feature extraction and information loss mitigation. Note that this use of RoIAlign does not contradict our earlier statement regarding its limitations in CMMB. Here, RoIAlign is adopted to improve the precision of attention score pooling, particularly in the case of small objects, not for enhancing the object detection model through contrastive learning. It thus acts as a natural and effective substitute for direct pooling. Specifically, we apply RoIAlign to extract proposal features $\mathcal{F}_{\mathrm{align}}^i$ from the corresponding query-fused encoder features $\mathcal{F}_{\mathrm{eq}}^i$:
\begin{equation}
\mathcal{F}_{\mathrm{align}}^i = \mathrm{RoIAlign}(\mathcal{F}_{\mathrm{eq}}^i, bbox),
\label{equ:roialign}
\end{equation}
where $\mathcal{F}_{\mathrm{align}}^i \in \mathbb{R}^{n_\mathrm{q} \times H_a \times W_a}$, with $H_a$ and $W_a$ representing the height and width of the RoIAlign output, respectively. Each query is then assigned an attention score of $H_a \times W_a$, enabling us to compute a set of objectness scores:
\begin{equation}
\mathcal{S} = \{\sum_{i = 3,4,5} \sum_{H_a \times W_a} \mathcal{F}_{\mathrm{align}}^i [j]\}_{j=0}^{n_\mathrm{q}-1},
\label{equ:objscore}
\end{equation}

The computed objectness scores $\mathcal{S}$ encode query-wise attention over multi-scale features. We observe that utilizing all object queries often leads to higher objectness attention for easily detectable foreground or background objects, aligning with findings in DETRDistill~\cite{chang2023detrdistill}. To mitigate foreground-background class imbalance and improve model discrimination for hard samples, we assign higher weights to foreground areas and hard samples with low attention scores. Specifically, we employ these normalized weights to refine the original Focal Loss~\cite{lin2017focal} in Deformable DETR~\cite{zhu2020deformable} for classification:
\begin{equation}
\mathcal{L}_{\mathrm{wcls}} = \frac{1}{n_\mathrm{q}} \sum_{i=0}^{n_\mathrm{q}-1} w_i \left[ - \alpha_t (1 - p_t)^\gamma \log(p_t) \right],
\label{equ:lwcls}
\end{equation}
and
\begin{equation}
w_i = (1 - \mathrm{MinMaxScaler}(\mathcal{S}[i])) ^ \beta,
\label{equ:wi}
\end{equation}
where $\mathrm{MinMaxScaler}(\mathcal{S}[i]) = \frac{\mathcal{S}[i] - \min(\mathcal{S})}{\max(\mathcal{S}) - \min(\mathcal{S})}$, and $\beta$ is a smoothing hyperparameter. $\alpha_t$ and $\gamma$ are the Focal Loss balancing parameters. By reweighting the query loss with attention-based objectness scores, our method encourages the model to focus more on foreground regions and hard samples. This enhances mutual learning under the teacher model’s pseudo supervision and improves model discriminability.

\begin{table*}[!t]
\centering
\caption{Results of cross-weather adaptation (foggy scenario). ``SF'' denotes Source-free. ``V'' and ``R'' in Backbone represent VGG and ResNet, respectively. ``FRCNN'' stands for Faster R-CNN. For each column, we bold the best and underline the second-best results separately for non-source-free and source-free approaches (excluding Oracle).}
\resizebox{0.95\textwidth}{!}{%
\begin{tabular}{lccc|cccccccc|c}
\toprule[1.5pt]
Method & SF & Backbone & Detector & Truck & Car & Rider & Person & Train & Motor & Bicycle & Bus & mAP \\
\midrule
Source & \usym{2613} & R-50 & DETR & 15.1 & 46.5 & 39.3 & 38.9 & 4.0 & 21.8 & 36.8 & 34.2 & 29.6 \\
\midrule
SIGMA++~\cite{li2023sigma++}& \usym{2613} & V-16 & \multirow{4}{*}{FCOS} & \underline{32.1} & 61.0 & 45.1 & 46.4 & 44.6 & 34.8 & 39.9 & 52.2 & 44.5 \\
CIGAR~\cite{liu2023cigar}& \usym{2613} & V-16 & & 27.8 & 62.1 & 47.3 & 46.1 & 44.3 & 33.7 & 41.3 & \textbf{56.6} & 44.9 \\
EPM~\cite{hsu2020every}& \usym{2613} & R-101 & & 29.4 & 57.1 & 43.6 & 41.5 & 39.7 & 29.0 & 36.1 & 44.9 & 40.2 \\
MGA-DA~\cite{zhou2022multi}& \usym{2613} & R-101 & & 30.2 & 61.5 & 47.3 & 43.1 & \textbf{50.3} & 27.9 & 36.9 & 53.2 & 43.8 \\
\midrule
SFA~\cite{wang2021exploring}& \usym{2613} & R-50 & \multirow{6}{*}{DETR} & 25.1 & 62.6 & 48.6 & 46.5 & 29.4 & 28.3 & 44.0 & 46.2 & 41.3 \\
MTTrans~\cite{yu2022mttrans}& \usym{2613} & R-50 & & 25.8 & 65.2 & 49.9 & 47.7 & 33.9 & 32.6 & 46.5 & 45.9 & 43.4 \\
DA-DETR~\cite{zhang2023detr}& \usym{2613} & R-50 & & 24.0 & 63.1 & 50.0 & 49.9 & 37.5 & 31.6 & 46.3 & 45.8 & 43.5 \\
MTM~\cite{weng2024mean}& \usym{2613} & R-50 & & \textbf{37.2} & 67.2 & 53.4 & 51.0 & 41.6 & 38.4 & 47.7 & 54.4 & 48.9 \\
BiADT~\cite{he2023bidirectional}& \usym{2613} & R-50 & & 31.7 & \underline{69.2} & \underline{58.9} & \underline{52.2} & \underline{45.1} & \underline{42.6} & \underline{51.3} & \underline{55.0} & \textbf{50.8} \\
ACCT~\cite{zeng2024enhancing}& \usym{2613} & R-50 & & 31.1 & \textbf{69.4} & \textbf{58.9} & \textbf{53.6} & 33.7 & \textbf{42.6} & \textbf{54.4} & 53.5 & \underline{49.6} \\
\midrule[1pt]
SED~\cite{li2021free}& \usym{1F5F8} & V-16 & \multirow{4}{*}{FRCNN} & 21.7 & 44.0 & 40.4 & 32.6 & 11.8 & 25.3 & 34.5 & 34.3 & 30.6 \\
IRG-SFDA~\cite{vs2023instance}& \usym{1F5F8} & R-50 & & 24.4 & 51.9 & 45.2 & 37.4 & 25.2 & 31.5 & 41.6 & 39.6 & 37.1 \\
AASFOD~\cite{chu2023adversarial}& \usym{1F5F8} & V-16 & & \underline{28.1} & 44.6 & 44.1 & 32.3 & 29.0 & 31.8 & 38.9 & 34.3 & 35.4 \\
BT~\cite{deng2024balanced}& \usym{1F5F8} & V-16 & & 24.3 & 52.7 & 47.1 & 38.4 & 36.3 & 30.2 & 40.1 & \underline{44.6} & 39.2 \\
\midrule
MT~\cite{tarvainen2017mean}& \usym{1F5F8} & R-50 & \multirow{4}{*}{DETR} & 23.5 & \underline{61.5} & 45.0 & 44.1 & 23.9 & 25.0 & 38.0 & 41.2 & 37.8 \\
TeST~\cite{sinha2023test}& \usym{1F5F8} & R-50 & & 27.5 & 54.8 & 46.2 & 45.1 & 11.9 & 32.0 & 44.1 & 44.4 & 38.2 \\
DRU~\cite{khanh2024dynamic}& \usym{1F5F8} & R-50 & & 26.2 & \textbf{62.5} & \textbf{51.5} & \textbf{48.3} & \underline{34.1} & \textbf{34.2} & \textbf{48.6} & 43.2 & \underline{43.6} \\

\textbf{FRANCK(ours)}& \usym{1F5F8} & R-50 & & \textbf{33.9} & 60.6 & \underline{49.3} & \underline{48.1} & \textbf{36.9} & \underline{34.0} & \underline{47.9} & \textbf{48.2} & \textbf{44.9} \\
\midrule
Oracle& \usym{1F5F8} & R-50 & DETR & 31.3 & 71.8 & 52.9 & 52.9 & 41.0 & 41.4 & 44.0 & 53.9 & 48.7 \\
\bottomrule[1.5pt]
\end{tabular}%
}
\label{tab:cw}
\end{table*}

\subsection{Uncertainty-weighted Query-fused Feature Distillation}
To ensure robust distillation between the teacher and student networks in SFOD, consistency regularization~\cite{li2023improving,vs2023instance,zhao2024multi} is a widely adopted approach. However, existing methods are primarily designed for Faster R-CNN, where shared proposals naturally align classification and localization scores for consistency regularization. In contrast, DETR employs a query-based mechanism, where different queries correspond to different objects, making direct consistency loss computation (\textit{e.g.}, the KL divergence) based on prediction indices infeasible. For instance, in most cases, the $i$-th predictions from the teacher and student networks correspond to different objects, rendering direct consistency enforcement impractical.

To address this challenge, we opt for feature imitation instead of logit mimicking for knowledge distillation, a strategy that has proven effective in object detection tasks~\cite{wang2019distilling,zhao2022decoupled,chang2023detrdistill}. Given that the student network in the original Mean Teacher framework processes strongly augmented images, we introduce an additional forward pass of these augmented images through the teacher model. This forward pass is solely used to extract image-level features for distillation. Following DETRDistill~\cite{chang2023detrdistill}, we reweight query-fused features based on prediction quality and formulate a unified feature distillation loss, denoted as $\mathcal{L}_{\mathrm{fdis}}$.

\noindent \textbf{Uncertainty-based Query Weighting}. We begin by describing our approach to extracting features for distillation. Utilizing object queries and image-level features, we construct objectness-weighted feature maps, similar to OSSR. However, DETRDistill~\cite{chang2023detrdistill} found that naïve objectness-weighted feature distillation is ineffective due to the varying contributions of queries. To address this, DETRDistill applies soft attention masks with quality scores~\cite{chen2021disentangle} derived from GT labels and teacher predictions. However, this does not apply to SFOD tasks, where GT labels are unavailable. To overcome this limitation, we leverage uncertainty estimation, using prediction entropy as the quality score. Specifically, we extract prediction scores from queries and compute entropy for teacher model predictions, denoted as $\mathcal{E} \in \mathbb{R}^{n_\mathrm{q}}$.
To assign higher weights to more reliable regions identified by the robust teacher model, we normalize the query weights using $\mathcal{W}_\mathrm{q} = (1 - \mathrm{MinMaxScaler}(\mathcal{E}))^{\beta^{'}}$, where we set $\beta^{'} = 1$ for simplicity.

\noindent \textbf{Query-fused Feature Distillation}. Following DETRDistill, we compute feature distillation loss by applying soft masks across queries and derive a unified weighted loss. Given a single query feature $\mathcal{F}_\mathrm{q}^j$ and the last encoder layer feature $\mathcal{F}_\mathrm{e}$, we derive the query-fused feature $\mathcal{F}_{\mathrm{eq}}^j \in \mathbb{R}^{H \times W}$ by:
\begin{equation}
\mathcal{F}_{\mathrm{eq}}^j = \mathcal{F}_\mathrm{e} \cdot (\mathcal{F}_\mathrm{q}^j)^T. 
\label{equ:feqj}
\end{equation}
Then by using $\mathcal{F}_{\mathrm{eq}}^j$ as an objectness-based soft mask, we can perform weighted feature imitation by:
\begin{equation}
\mathcal{L}_{\mathrm{fdis}} = \frac{1}{n_\mathrm{q}HWC} \sum_{j=0}^{n_\mathrm{q}-1} \mathcal{W}_\mathrm{q}^j \left\| \mathcal{F}_{\mathrm{eq}}^j \odot (\mathcal{F}_\mathcal{T} - \mathcal{F}_\mathcal{S}) \right\|_2^2,
\label{equ:lfdis}
\end{equation}
where $\mathcal{W}_\mathrm{q}^j$ and $\mathcal{F}_{\mathrm{eq}}^j$ denote the $j$-th elements of $\mathcal{W}_\mathrm{q}$ and the query-fused feature, respectively. $\mathcal{F}_\mathcal{T}$ and $\mathcal{F}_\mathcal{S}$ represent the last encoder layer features of the teacher and student models, respectively, and $\odot$ denotes Hadamard product. By incorporating weighted feature distillation, the teacher guides the student to generate more stable feature representations, improving the robustness of knowledge transfer.

\subsection{Dynamic Teacher Updating Interval}
To enhance the conventional Mean Teacher self-training approach, we introduce an improved updating mechanism termed Dynamic Teacher Updating Interval (DTUI). As discussed in Sec.~\ref{sec:preliminaries}, we adopt the Mean Teacher framework to ensure a robust and efficient adaptation scheme. Specifically, after performing forward propagation on both the teacher and student networks, we first filter pseudo-labels by applying a confidence threshold of 0.3 to the teacher’s predictions. We then compute the total loss function as follows:
\begin{equation}
\mathcal{L}_{total} = \mathcal{L}_{\mathrm{wcls}} + \mathcal{L}_{\mathrm{reg}} + \mathcal{L}_{\mathrm{aux}} + \omega_1\mathcal{L}_{\mathrm{cont}} + \omega_2\mathcal{L}_{\mathrm{fdis}},
\label{equ:ltotal}
\end{equation}
where $\mathcal{L}_{\mathrm{wcls}}$, $\mathcal{L}_{\mathrm{reg}}$, and $\mathcal{L}_{\mathrm{aux}}$ collectively form the detection loss $\mathcal{L}_{\mathrm{det}}$ by replacing the original classification loss with Eq.~(~\ref{equ:lwcls}). The student model is then optimized using $\mathcal{L}_{total}$, while the teacher model is updated using EMA with a momentum factor of $\alpha_{\mathrm{EMA}}$.

\noindent \textbf{DTUI.} In SFOD tasks, domain shifts and the absence of ground truth labels can destabilize the student model’s prediction and optimization, leading to biased teacher updates and reduced mutual learning effectiveness~\cite{zhao2023towards,zhao2024multi}. To mitigate this issue, AASFOD~\cite{chu2023adversarial} and DACA~\cite{zhao2024multi} adopt fixed EMA update intervals for each experiment, which is helpful but ignores the adaptation progress over time. In contrast, we propose a dynamic EMA interval $i_{\mathrm{EMA}}$, formulated as:
\begin{equation}
i_{\mathrm{EMA}} = \delta + \left\lfloor e / \epsilon \right\rfloor,
\label{equ:iema}
\end{equation}
where $\delta$ is a base interval that controls the steps for stable knowledge accumulation, $e$ represents the current epoch index, and $\epsilon$ denotes the increment rate. Under this dynamic strategy, while the student network is updated at every iteration using Eq.~(~\ref{eq:student_update}), the teacher network is updated every $i_{\mathrm{EMA}}$ iterations. This linear interval adjustment allows for frequent parameter exploration at the early stages of adaptation, facilitating a more effective search of the parameter space, and progressively stabilizing the model’s updates as training progresses.

\begin{table*}[!t]
\centering
\caption{Results of cross-weather adaptation (rainy scenario). ``SF'' denotes Source-free. ``V'' and ``R'' in Backbone represent VGG and ResNet, respectively. ``FRCNN'' stands for Faster R-CNN. For each column, we bold the best and underline the second-best results separately for non-source-free and source-free approaches (excluding Oracle).}
\label{tab:c2r}
\resizebox{0.95\textwidth}{!}{%
\begin{tabular}{lccc|cccccccc|c}
\toprule[1.5pt]
Method & SF & Backbone & Detector & Truck & Car & Rider & Person & Train & Motor & Bicycle & Bus & mAP \\
\midrule
Source& \usym{2613} & R-50 & DETR & 35.0 & 69.0 & 47.4 & 50.9 & 31.5 & 30.5 & 39.9 & 51.0 & 44.4 \\
\midrule
DA-Detect~\cite{li2023domain2}& \usym{2613} & R-50 & FRCNN & 38.8 & 61.7 & 47.0 & 40.2 & \textbf{47.2} & 34.4 & 29.0 & \underline{59.7} & 46.0 \\
\midrule
BiADT~\cite{he2023bidirectional}& \usym{2613} & R-50 & \multirow{3}{*}{DETR} & \textbf{41.9} & \textbf{75.2} & \textbf{52.8} & \textbf{54.8} & \underline{42.6} & \underline{36.3} & \textbf{47.3} & \textbf{61.5} & \textbf{51.5} \\
AQT~\cite{huang2022aqt}& \usym{2613} & R-50 & & 28.5 & 68.1 & 45.8 & 46.6 & 29.5 & 34.2 & 44.2 & 50.2 & 43.4 \\
SFA~\cite{wang2021exploring}& \usym{2613} & R-50 & & \underline{40.4} & \underline{73.3} & \underline{49.5} & \underline{52.5} & 40.1 & \textbf{36.6} & \underline{44.2} & 52.5 & \underline{48.6} \\
\midrule[1pt]
IRG-SFDA~\cite{vs2023instance}& \usym{1F5F8} & R-50 & FRCNN & 33.1 & 60.4 & 46.7 & 39.2 & 33.1 & 34.4 & 42.2 & 52.1 & 42.7 \\
\midrule
MT~\cite{tarvainen2017mean}& \usym{1F5F8} & R-50 & \multirow{3}{*}{DETR} & 34.2 & 73.3 & 49.6 & 51.0 & 39.0 & 29.2 & 45.1 & 48.0 & 46.2 \\
DRU~\cite{khanh2024dynamic}& \usym{1F5F8} & R-50 & & \underline{35.3} & \underline{73.7} & \textbf{51.7} & \underline{52.5} & \underline{39.4} & \underline{32.7} & \underline{45.8} & \underline{53.7} & \underline{48.1} \\

\textbf{FRANCK(ours)}& \usym{1F5F8} & R-50 & & {\textbf{41.1}} & \textbf{74.3} & \underline{50.9} & \textbf{52.8} & \textbf{45.1} & \textbf{33.7} & \textbf{47.4} & \textbf{54.7} & \textbf{50.0} \\
\midrule
Oracle& \usym{1F5F8} & R-50 & DETR & 42.7 & 73.2 & 53.4 & 53.2 & 48.0 & 38.6 & 46.7 & 55.8 & 51.5 \\
\bottomrule[1.5pt]
\end{tabular}%
}
\end{table*}

\begin{table}[!t]
\centering\footnotesize
\caption{Results of Synthetic-to-real (S2R) and Cross-dataset (K2C) adaptation. ``SF'' denotes Source-free. ``V'' and ``R'' in Backbone represent VGG and ResNet, respectively. ``FRCNN'' stands for Faster R-CNN. For each column, we bold the best and underline the second-best results separately for non-source-free and source-free approaches (excluding Oracle).}
\resizebox{0.95\linewidth}{!}{%
\begin{tabular}{lccc|cc}
\toprule[1.5pt]
\multirow{2}{*}{Method} & \multirow{2}{*}{SF} & \multirow{2}{*}{Backbone} & \multirow{2}{*}{Detector} & \multicolumn{2}{c}{mAP} \\
\cmidrule(lr){5-6}
 &  &  &  & S2R & K2C \\
\midrule
Source & \usym{2613} & R-50 & DETR & 50.8 & 33.9 \\
\midrule
SFA~\cite{wang2021exploring} & \usym{2613} & R-50 & DETR & \underline{52.6} & \underline{46.7} \\
DA-DETR~\cite{zhang2023detr} & \usym{2613} & R-50 & DETR & \textbf{54.7} & \textbf{48.9} \\
\midrule[1pt]
IRG-SFDA~\cite{vs2023instance} & \usym{1F5F8} & R-50 & FRCNN & 46.9 & \underline{45.2} \\
AASFOD~\cite{chu2023adversarial} & \usym{1F5F8} & V-16 & FRCNN & 44.9 & 44.0 \\
\midrule
MT~\cite{tarvainen2017mean} & \usym{1F5F8} & R-50 & DETR & 57.0 & 42.5 \\
TeST~\cite{sinha2023test} & \usym{1F5F8} & R-50 & DETR & 57.9 & 42.8 \\
DRU~\cite{khanh2024dynamic} & \usym{1F5F8} & R-50 & DETR & \underline{58.7} & 45.1 \\
\textbf{FRANCK(ours)} & \usym{1F5F8} & R-50 & DETR & \textbf{63.1} & \textbf{48.5} \\
\midrule
Oracle & \usym{1F5F8} & R-50 & DETR & 75.9 & 75.9 \\
\bottomrule[1.5pt]
\end{tabular}%
}
\label{tab:s2r-k2c}
\end{table}

\section{Experiments} \label{sec:experiments}
This section presents the datasets, experimental setup, experimental results, and comprehensive analyses from quantitative, ablation, and visualization studies to validate our method.
\subsection{Datasets}
We evaluate our approach on four widely used object detection datasets: Cityscapes~\cite{cordts2016cityscapes}, Foggy Cityscapes~\cite{sakaridis2018semantic}, Sim10k~\cite{johnson2016driving}, and BDD100K~\cite{yu2018bdd100k}, along with a synthesized rainy Cityscapes dataset. Our experiments cover cross-weather, synthetic-to-real, and cross-scene adaptation.

\noindent \textbf{Cross-weather Adaptation.} Cityscapes~\cite{cordts2016cityscapes} is an urban scene dataset with 2,975 training and 500 validation images from various cities. Foggy Cityscapes~\cite{sakaridis2018semantic} extends it via synthetic fog generation. We use Cityscapes and its foggy variant (with 0.02 fog density) as the source and target domains, respectively. To further evaluate robustness under adverse weather, we introduce a rainy version of Cityscapes using RainMix~\cite{guo2021efficientderain}, following prior works~\cite{li2023domain,li2023domain2}, enabling assessment across diverse weather conditions~\cite{hu2019depth,li2023domain2,sindagi2020prior}.

\noindent \textbf{Synthetic-to-real Adaptation.} Sim10k~\cite{johnson2016driving}, generated from the GTA V game, contains 9,000 training and 1,000 validation images. In this setting, Sim10k is the source domain and Cityscapes is the target domain. This setting assesses the ability to generalize from synthetic data to real data distributions, offering benefits such as reduced data collection costs and enhanced data diversity in real-world scenarios.

\noindent \textbf{Cross-scene Adaptation.} BDD100K~\cite{yu2018bdd100k} is a large-scale autonomous driving dataset covering different times of the day. Following prior work~\cite{wang2021exploring,yu2022mttrans,khanh2024dynamic}, we use only daytime images, comprising 36,728 training and 5,258 validation images. In this setting, Cityscapes is used as the source domain and BDD100K daytime as the target domain. This assesses a detection model’s adaptability across diverse scenes.

\noindent \textbf{Cross-dataset Adaptation.} KITTI~\cite{geiger2012we} is an autonomous driving dataset collected from diverse real-world scenes. In our setting, all 7,481 annotated images from KITTI are used as the source domain, while Cityscapes serves as the target domain. This setup evaluates the detection model’s ability to adapt across different camera systems and dataset characteristics.

\noindent \textbf{Cityscapes-to-ACDC Adaptation.} ACDC~\cite{sakaridis2021acdc} is a dataset designed for comprehensive understanding of autonomous driving scenes. It encompasses four types of challenging real‑world weather conditions, including snow, rain, night, and fog. This adaptation setting is used to further evaluate the effectiveness and robustness of our method under diverse and complex real‑world domain shifts.


\subsection{Baselines}
We compare our method against multiple baseline settings, including source-only, DAOD, SFOD, and Oracle.

\noindent \textbf{Source-only.} For Source-only baselines, the source pre-trained model is directly evaluated on the target domain without adaptation, serving as a lower bound for domain adaptation.

\noindent \textbf{DAOD.} We compare our approach with previous DETR-based DAOD methods, including SFA~\cite{wang2021exploring}, MTTrans~\cite{yu2022mttrans}, DA-DETR~\cite{zhang2023detr}, MTM~\cite{weng2024mean}, BiADT~\cite{he2023bidirectional}, and ACCT~\cite{zeng2024enhancing}. Among them, only BiADT uses DAB-Deformable-DETR~\cite{liu2022dab}, a variant of Deformable DETR~\cite{zhu2020deformable}, as its base detector, while all others adopt Deformable DETR. Additionally, we also compare with approaches based on different detectors, including SIGMA++\cite{li2023sigma++}, CIGAR\cite{liu2023cigar}, EPM~\cite{hsu2020every}, and MGA-DA~\cite{zhou2022multi}. These comparisons provide insights into DAOD performance on DETRs and highlight our method’s advantages.

\begin{table*}[!t]
  \centering
  \caption{Results of Cross-scene adaptation. ``SF'' denotes Source-free. ``V'' and ``R'' in Backbone represent VGG and ResNet, respectively. ``FRCNN'' stands for Faster R-CNN. For each column, we bold the best and underline the second-best results separately for non-source-free and source-free approaches (excluding Oracle).}
    \resizebox{0.92\linewidth}{!}{%
    \begin{tabular}{lccc|ccccccc|c}
    \toprule[1.5pt]
    Method & SF & Backbone & Detector & Truck & Car & Rider & Person & Motor & Bicycle & Bus & mAP \\
    \midrule
    Source& \usym{2613} & R-50   & DETR & 17.5  & 57.0  & 29.4  & 43.7  & 15.6  & 17.7  & 17.6  & 28.3 \\
    \midrule
    SIGMA++~\cite{li2023sigma++}& \usym{2613} & V-16   & \multirow{2}[2]{*}{FCOS} & 21.1  & \underline{65.6}  & 30.4  & 47.5  & 27.1  & 17.8  & 26.3  & 33.7  \\
          EPM~\cite{hsu2020every}& \usym{2613}   & V-16   &       & 18.8  & 55.8  & 26.8  & 39.6  & 20.1  & 14.5  & 19.1  & 27.8  \\
\midrule        SFA~\cite{wang2021exploring}& \usym{2613}   & R-50   & \multirow{5}[2]{*}{DETR} & 19.1  & 57.5  & 27.6  & 40.2  & 19.2  & 15.4  & 23.4  & 28.9  \\
          MTTrans~\cite{yu2022mttrans}& \usym{2613} & R-50   &       & \underline{25.1}  & 61.5  & 30.1  & 44.1  & 23.0  & 17.7  & \underline{26.9}  & 32.6  \\
          MTM~\cite{weng2024mean}& \usym{2613}   & R-50   &       & 23.0  & \textbf{68.8}  & \underline{35.1}  & \textbf{53.7}  & \underline{28.0}  & \underline{23.8}  & \textbf{28.8}  & \underline{37.3}  \\
          BiADT~\cite{he2023bidirectional}& \usym{2613} & R-50   &       & 17.4  & 60.9  & 34.0  & 42.1  & 25.7  & 18.2  & 19.5  & 31.1  \\
          ACCT~\cite{zeng2024enhancing}& \usym{2613}  & R-50   &       & \textbf{26.0}  & 61.8  & \textbf{41.4}  & \underline{51.8}  & \textbf{36.9}  & \textbf{31.7}  & 23.4  & \textbf{39.0} \\
    \midrule[1pt]
    SED~\cite{li2021free}& \usym{1F5F8}   & V-16   & \multirow{3}[2]{*}{FRCNN} & 20.6  & 50.4  & 32.6  & 32.4  & 25.0  & 18.9  & 23.4  & 29.0  \\
          AASFOD~\cite{chu2023adversarial}& \usym{1F5F8} & V-16   &       & 26.6  & 50.2  & 36.3  & 33.2  & 22.5  & \textbf{28.2}  & 24.4  & 31.6  \\
          BT~\cite{deng2024balanced}& \usym{1F5F8}    & V-16   &       & 24.2  & 50.4  & 34.6  & 32.7  & 28.5  & 24.7  & 24.9  & 31.4  \\
\midrule         MT~\cite{tarvainen2017mean}& \usym{1F5F8}   & R-50   & \multirow{4}[2]{*}{DETR} & 21.8  & 63.0  & 36.0  & 50.5  & 20.9  & 22.0  & 22.4  & 33.8  \\
          TeST~\cite{sinha2023test}& \usym{1F5F8}  & R-50   &       & 23.2  & \underline{63.8}  & \underline{37.5}  & \underline{51.7}  & 22.7  & 23.4  & 23.6  & 35.1  \\
          DRU~\cite{khanh2024dynamic}& \usym{1F5F8}   & R-50   &       & \underline{27.1}  & 62.7  & 36.9  & 45.8  & \textbf{32.5}  & 22.7  & \underline{28.1}  & \underline{36.6}  \\
         \textbf{FRANCK(ours)}& \usym{1F5F8}  & R-50   &       & \textbf{29.5}  & \textbf{65.6}  & \textbf{43.3}  & \textbf{55.0}  & \underline{30.0}  & \underline{28.0}  & \textbf{33.6}  & \textbf{40.7} \\
    \midrule
    Oracle& \usym{1F5F8} & R-50 & DETR & 66.9 & 87.9 & 56.4 & 74.9 & 53.8 & 68.3 & 55.0 & 66.2  \\
    \bottomrule[1.5pt]
    \end{tabular}%
   }
  \label{tab:cs}%
\end{table*}%

\begin{table}[!t]
\centering
\caption{Results of Cityscapes-to-ACDC adaptation on four diverse weather conditions. ``SF'' denotes Source-free. ``FRCNN'' stands for Faster R-CNN. We bold the best and underline the second-best results for source-free approaches.}
\label{tab:city_to_acdc}
\resizebox{0.95\linewidth}{!}{%
\begin{tabular}{l|c|l|cccc}
\toprule[1.5pt]
Method & SF & Detector & Snow & Rain & Night & Fog \\
\midrule
AT~\cite{li2022at}           & \usym{2613} & \multirow{2}{*}{FRCNN} & 55.2 & 37.7 & 29.5 & 62.2 \\
DT-G~\cite{lavoie2025large}     & \usym{2613} &  & 56.8 & 39.0 & 36.4 & 68.6 \\
\midrule
DRU~\cite{khanh2024dynamic}     & \usym{1F5F8} & \multirow{2}{*}{DETR} & \underline{37.9} & \underline{26.3} & \underline{16.5} & \underline{45.4} \\
\textbf{FRANCK(ours)}           & \usym{1F5F8} &  & \textbf{42.4} & \textbf{29.9} & \textbf{17.5} & \textbf{49.0} \\
\bottomrule[1.5pt]
\end{tabular}%
}
\end{table}


\noindent \textbf{SFOD.} We include DETR-based SFOD methods such as TeST~\cite{sinha2023test} and the concurrent work DRU~\cite{khanh2024dynamic}. Since TeST is not open-source, we implement and reproduce it, tuning hyperparameters for optimal results. OnDA-DETR~\cite{suzuki2023onda} is omitted due to its similarity to vanilla Mean Teacher, for which we conduct separate experiments. Given the limited number of DETR-based SFOD works, we also compare against Faster R-CNN-based SFOD methods, including SED~\cite{li2021free}, IRG-SFDA~\cite{vs2023instance}, AASFOD~\cite{chu2023adversarial}, and BT~\cite{deng2024balanced}.

\noindent \textbf{Oracle.} For Oracle baselines, models are trained and tested directly on the labeled target domain without source pretrain, representing the upper bound of DAOD to some extent.

Following prior work, detection performance is evaluated using mean average precision (mAP) with IoU$=$0.5. From Tab.~\ref{tab:cw} to Tab.~\ref{tab:s2r-k2c}, we use R-50 and R-101 to refer to ResNet-50 and ResNet-101\cite{he2016deep}, respectively, and V-16 to denote the VGG-16\cite{simonyan2014very} network. Our comparisons include methods based on FCOS\cite{tian2019fcos}, DETR\cite{carion2020end, zhu2020deformable}, and Faster R-CNN (FRCNN)\cite{ren2015faster}. Notably, apart from BiADT\cite{he2023bidirectional}, which adopts DAB-Deformable DETR\cite{liu2022dab}, which is a variant of Deformable DETR\cite{zhu2020deformable}, all DETR-based approaches in our study are built upon Deformable DETR~\cite{zhu2020deformable}.

\subsection{Implementation Details}
In this section, we provide the implementation details of the experiments. The hyperparameters, along with their notations, descriptions, and values, are summarized in Tab.~\ref{tab:hyperparam}. In the source pretraining stage, we train the model for 50 epochs, starting with a learning rate of $2 \times 10^{-4}$, which is reduced by a factor of 0.1 after 40 epochs. During the adaptation stage, the model is trained for 30 epochs with a fixed learning rate of $5 \times 10^{-5}$. All experiments use a batch size of 2 per GPU, with training conducted on 4 NVIDIA RTX 4090 GPUs.

\subsection{Comparison with State-of-the-art}
In this section, we present quantitative comparisons with state-of-the-art methods. The results are summarized in Tab.~\ref{tab:cw}, Tab.~\ref{tab:c2r}, Tab.~\ref{tab:s2r-k2c}, Tab.~\ref{tab:cs}, and Tab.~\ref{tab:city_to_acdc}, with the best mAP(\%) values and second-best mAP values (excluding Oracle results) bolded and underlined, respectively. Our method achieves state‑of‑the‑art results across multiple domain adaptation settings. In cross‑weather adaptation, it attains 44.9 mAP on Cityscapes to Foggy Cityscapes, showing that the model remains robust and effectively overcomes domain shift even when object visibility is severely reduced, and reaches 50 mAP for rainy conditions. In synthetic‑to‑real adaptation, it achieves 63.1 mAP, demonstrating strong transferability from cost‑effective synthetic data to complex real scenes. In Cityscapes‑to‑ACDC adaptation, it surpasses DRU across all weather conditions with gains of 1.0 to 4.6 mAP, confirming its ability to handle diverse environments. In cross‑scene and cross‑dataset adaptation, it reaches 40.7 mAP and 48.5 mAP respectively, further proving its robustness across varying locations and data sources.

\begin{table*}[!t]
  \centering
  \caption{Performance Comparison of Different Backbones. We use Deformable DETR as the base detector. We bold the best and underline the second-best results separately for each backbone.}
  \label{tab:backbones}
  \resizebox{0.9\linewidth}{!}{
  \begin{tabular}{l|c|cccccccc|c}
    \toprule[1.5pt]
    Method & Backbone & Truck & Car & Rider & Person & Train & Motor & Bicycle & Bus & mAP \\
    \midrule
    Source & \multirow{4}{*}{R-50~\cite{he2016deep}} & 15.1  & 46.5  & 39.3  & 38.9  & 4.0   & 21.8  & 36.8  & 34.2  & 29.6  \\ 
    MT~\cite{tarvainen2017mean}     &  & 23.5  & \underline{61.5}  & 45.0  & 44.1  & 23.9  & 25.0  & 38.0  & 41.2  & 37.8  \\ 
    DRU~\cite{khanh2024dynamic}    &  & \underline{26.2}  & \textbf{62.5} & \textbf{51.5} & \textbf{48.3} & \underline{34.1}  & \textbf{34.2} & \textbf{48.6} & \underline{43.2}  & \underline{43.6}  \\ 
     \textbf{FRANCK(ours)} &     & \textbf{33.9} & 60.6  & \underline{49.3}  & \underline{48.1}  & \textbf{36.9} & \underline{34.0}  & \underline{47.9}  & \textbf{48.2} & \textbf{44.9} \\ 
    \midrule
    Source & \multirow{4}{*}{ViT-B~\cite{dosovitskiy2020image}} & 17.8  & 46.8  & 41.3  & 38.2  & 3.7   & 24.2  & 37.2  & 30.1  & 29.9  \\ 
    MT~\cite{tarvainen2017mean}     &  & 22.3  & 62.0  & 46.1  & \underline{45.9}  & 11.2  & \underline{27.5}  & 41.1  & 37.7  & 36.7  \\ 
    DRU~\cite{khanh2024dynamic}    &  & \underline{22.5}  & \textbf{62.6} & 45.9  & 44.8  & \textbf{20.9} & 25.5  & \underline{41.1}  & \underline{39.2}  & \underline{37.8}  \\ 
     \textbf{FRANCK(ours)} &     & \textbf{25.7} & \underline{62.2}  & \textbf{49.6} & \textbf{46.7} & \underline{14.1}  & \textbf{30.4} & \textbf{42.0} & \textbf{43.3} & \textbf{39.3} \\ 
    \midrule
    Source & \multirow{4}{*}{Swin-T~\cite{liu2021swin}} & 30.0  & 55.5  & 48.5  & 45.0  & 15.8  & 35.3  & 42.6  & 41.3  & 39.2  \\ 
    MT~\cite{tarvainen2017mean}     &  & 33.3  & \underline{65.5}  & 48.9  & 47.2  & 19.7  & 34.4  & 46.9  & 40.9  & 42.1  \\ 
    DRU~\cite{khanh2024dynamic}    &  & \underline{35.0}  & \textbf{66.4} & \underline{49.7}  & \underline{49.8}  & \textbf{30.4} & \underline{38.1}  & \underline{47.6}  & \underline{46.9}  & \underline{45.5}  \\ 
     \textbf{FRANCK(ours)} &     & \textbf{35.2} & 65.0  & \textbf{52.3} & \textbf{50.5} & \underline{28.4}  & \textbf{41.0} & \textbf{50.0} & \textbf{47.2} & \textbf{46.2} \\ 
    \midrule
    Source & \multirow{4}{*}{Swin-S~\cite{liu2021swin}} & 29.9  & 56.2  & 50.5  & 45.6  & \textbf{35.9}  & 36.6  & 42.7  & 46.6  & 43.0  \\ 
    MT~\cite{tarvainen2017mean}     &  & 36.6  & \textbf{66.9} & 52.1  & 49.1  & 25.3  & 40.2  & 47.2  & 47.0  & 45.6  \\ 
    DRU~\cite{khanh2024dynamic}    &  & \textbf{37.8} & 65.3  & \underline{53.8}  & \underline{50.8}  & \underline{27.6} & \underline{41.8}  & \underline{49.2}  & \underline{50.7}  & \underline{47.1}  \\ 
     \textbf{FRANCK(ours)} &     & \underline{37.3}  & \underline{65.8}  & \textbf{54.0} & \textbf{51.4} & 26.9  & \textbf{42.3} & \textbf{50.1} & \textbf{52.1} & \textbf{47.5} \\ 
    \midrule
    Source & \multirow{4}{*}{Swin-B~\cite{liu2021swin}} & 37.7  & 62.5  & 51.3  & 49.3  & 39.2  & 38.6  & 45.2  & 46.2  & 46.2  \\ 
    MT~\cite{tarvainen2017mean}     &  & \underline{41.2}  & 67.2  & 48.6  & 46.4  & 39.0  & \underline{43.2}  & 46.8  & 49.7  & 47.9  \\ 
    DRU~\cite{khanh2024dynamic}    &  & 39.4  & \underline{69.1}  & \underline{52.5}  & \underline{51.8}  & \underline{42.1} & \textbf{44.7} & \underline{50.8}  & \underline{56.4}  & \underline{50.9}  \\ 
     \textbf{FRANCK(ours)} &     & \textbf{43.1} &\textbf{69.2} & \textbf{53.4} & \textbf{53.1} & \textbf{42.1} & 42.7  & \textbf{52.6} & \textbf{56.8} & \textbf{51.6} \\ 
    \midrule
    Source & \multirow{4}{*}{Swin-L~\cite{liu2021swin}} & 39.4  & 63.5  & 49.5  & 49.2  & 48.9  & 41.9  & 45.6  & 52.7  & 48.9  \\ 
    MT~\cite{tarvainen2017mean}     &  & 44.0  & 68.0  & 51.5  & 49.7  & 42.2  & 43.6  & 47.3  & 55.2  & 49.5  \\ 
    DRU~\cite{khanh2024dynamic}    &  & \underline{45.0}  & \underline{68.8}  & \underline{52.9}  & \underline{53.1}  & \textbf{47.8} & \underline{45.2} & \underline{50.6}  & \underline{56.6}  & \underline{52.5}  \\ 
     \textbf{FRANCK(ours)} &     & \textbf{46.0} & \textbf{69.4} & \textbf{55.8} & \textbf{54.4} & \underline{46.1} & \textbf{48.2}  & \textbf{51.2} & \textbf{59.1} & \textbf{53.8} \\ 
    \bottomrule[1.5pt]
  \end{tabular}
 }
\end{table*}

\begin{table*}[!t] 
  \centering
  \caption{Performance Comparison of Different DETR Variants. We use Resnet-50 as the backbone. We bold the best and underline the second-best results separately for each kind of detector.}
  \label{tab:variants}
  \resizebox{0.95\linewidth}{!}{

    \begin{tabular}{l|c|cccccccc|c}
    \toprule[1.5pt]
    Method & Detector & Truck & Car & Rider & Person & Train & Motor & Bicycle & Bus  & mAP \\
    \midrule
    Source & \multirow{3}[2]{*}{Deformable DETR~\cite{zhu2020deformable}} & 15.1  & 46.5  & 39.3  & 38.9  & 4.0   & 21.8  & 36.8  & 34.2  & 29.6  \\
    MT~\cite{tarvainen2017mean}    &       & \underline{23.5}  & \textbf{61.5} & \underline{45.0}  & \underline{44.1}  & \underline{23.9}  & \underline{25.0}  & \underline{38.0}  & \underline{41.2}  & \underline{37.8}  \\
     \textbf{FRANCK(ours)} &          & \textbf{33.9} & \underline{60.6}  & \textbf{49.3} & \textbf{48.1} & \textbf{36.9} & \textbf{34.0} & \textbf{47.9} & \textbf{48.2} & \textbf{44.9} \\
    \midrule
    Source & \multirow{3}[2]{*}{DINO DETR~\cite{zhang2022dino}} & \underline{18.9}  & 46.1  & 42.0  & 39.5  & 7.5   & 24.1  & 40.1  & 35.1  & 31.7  \\
    MT~\cite{tarvainen2017mean}    &       & 18.5  & \underline{64.7}  & \underline{46.0}  & \underline{49.5}  & \underline{37.8}  & \underline{35.6}  & \underline{44.8}  & \underline{44.7}  & \underline{42.7}  \\
     \textbf{FRANCK(ours)} &          & \textbf{29.2} & \textbf{65.0} & \textbf{54.3} & \textbf{51.8} & \textbf{40.3} & \textbf{35.6} & \textbf{50.6} & \textbf{49.7} & \textbf{47.1} \\
    \midrule
    Source & \multirow{3}[2]{*}{RT DETR~\cite{zhao2024detrs}} & 29.9  & 54.0  & 46.8  & 38.7  & 21.9  & 29.4  & 39.3  & 43.7  & 38.0  \\
    MT~\cite{tarvainen2017mean}    &       & \underline{34.5}  & \underline{64.5} & \underline{51.2}  & \underline{45.0}  & \underline{37.4}  & \underline{36.3}  & \underline{43.4}  & \underline{54.6}  & \underline{45.9}  \\
     \textbf{FRANCK(ours)} &          & \textbf{44.2} & \textbf{73.4} & \textbf{54.8} & \textbf{51.3} & \textbf{52.1}  & \textbf{42.9} & \textbf{48.6} & \textbf{65.8} & \textbf{54.1} \\
    \bottomrule[1.5pt]
    \end{tabular}%

 }
\end{table*}




\subsection{Ablation Study} \label{ssec:abl}
In this section, we present several ablation studies to assess the design and effectiveness of our method. Unless specified, we conduct all ablation studies on Cityscapes to Foggy Cityscapes adaptation in the Cross-weather setting.

\noindent \textbf{Component Ablation.} We evaluate the impact of different components by comparing performance with and without each component. As shown in Tab.~\ref{tab:abl_components}, the ablation results demonstrate that (1) the Mean Teacher strategy and dynamic MT updating interval improve SFOD performance by enhancing robust training and knowledge distillation, and (2) The proposed methods, including CMMB, OSSR, and UQFD, each contribute to the final detection performance, resulting in a 3.2 mAP gain based on Mean Teacher with DTUI with only 28.5\% additional training time.

\noindent \textbf{Effectiveness Across Backbones.}
While our main experiments use ResNet‑50, we further evaluate transformer‑based backbones, including ViT‑Base~\cite{dosovitskiy2020image} and Swin Transformers~\cite{liu2021swin} of different scales—Swin‑T (tiny), Swin‑S (small), Swin‑B (base), and Swin‑L (large). As shown in Tab.~\ref{tab:backbones}, vanilla ViTs perform worse overall due to their single‑scale, low‑resolution features~\cite{wang2021pyramid}, yet the trends remain consistent: (1) domain shifts lead to similar performance degradation across all backbones, and SFOD methods effectively mitigate this issue; and (2) our method consistently surpasses MT~\cite{tarvainen2017mean} and DRU~\cite{khanh2024dynamic} on every backbone tested, highlighting both its robustness and its ability to generalize across different feature extractors.

\begin{figure*}[thb] \centering
   
    \includegraphics[width=0.15\textwidth]{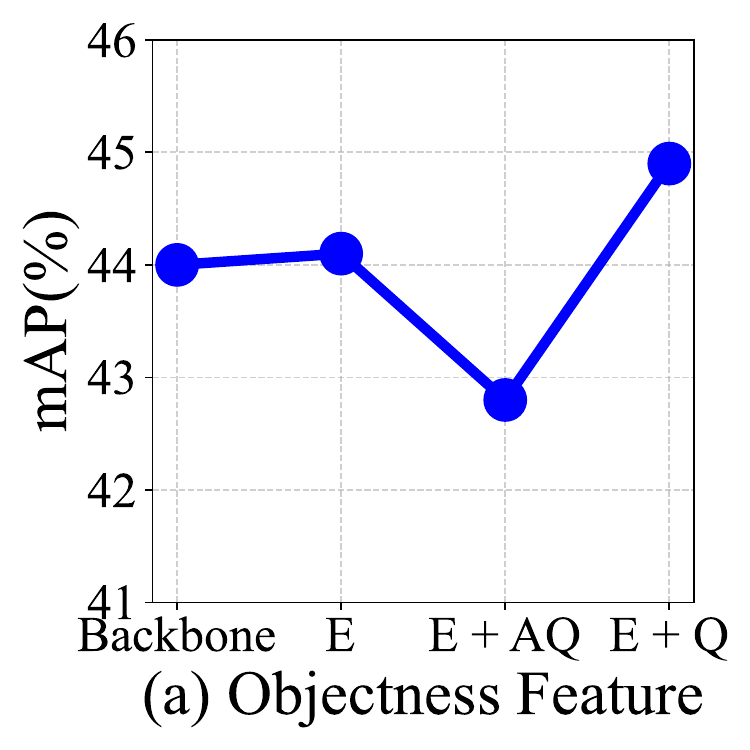}
    \includegraphics[width=0.15\textwidth]{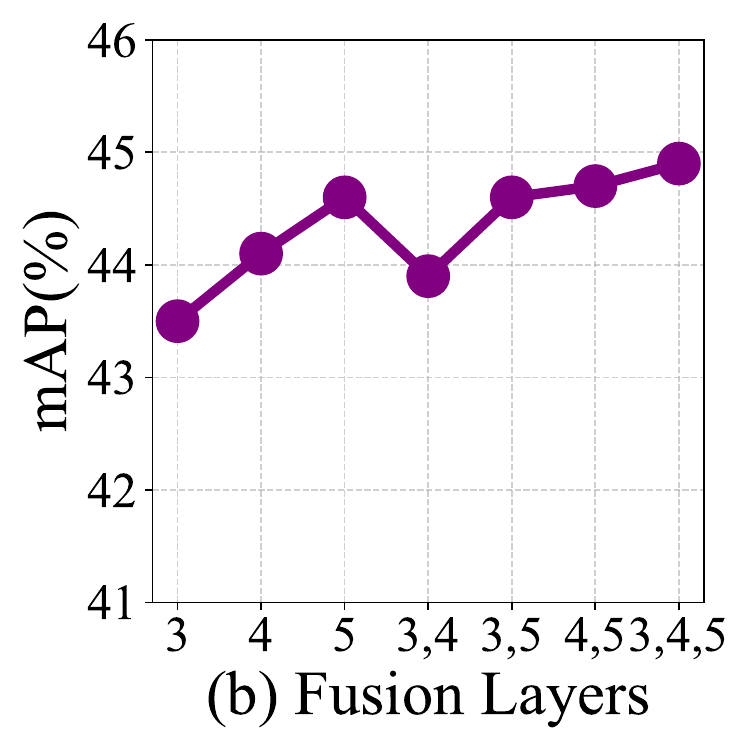}
    \includegraphics[width=0.15\textwidth]{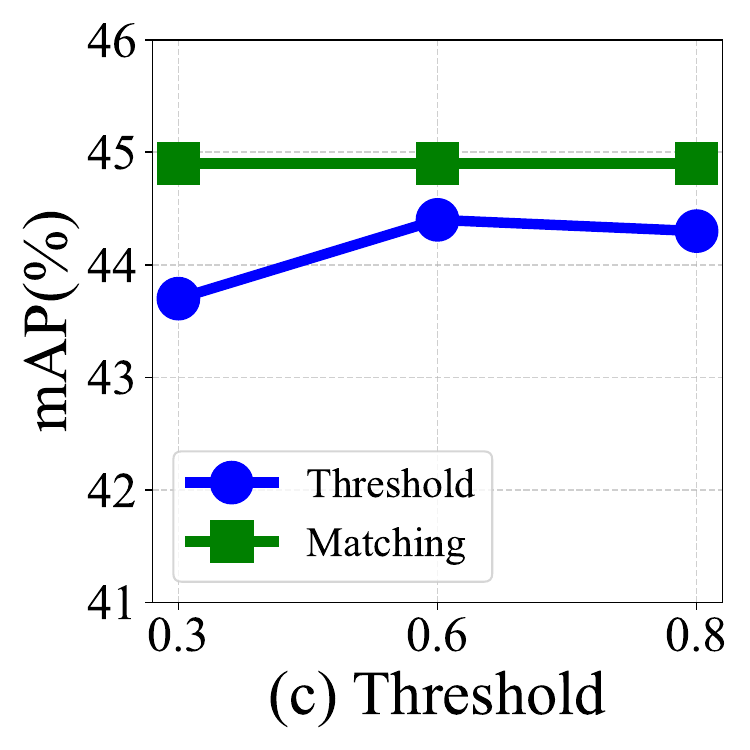}
    \includegraphics[width=0.15\textwidth]{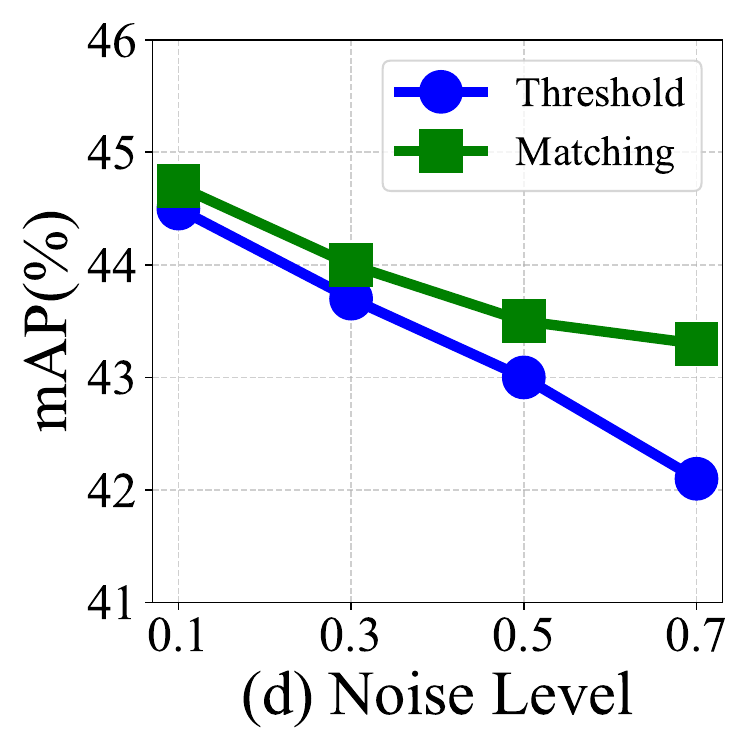}
    \includegraphics[width=0.15\textwidth]{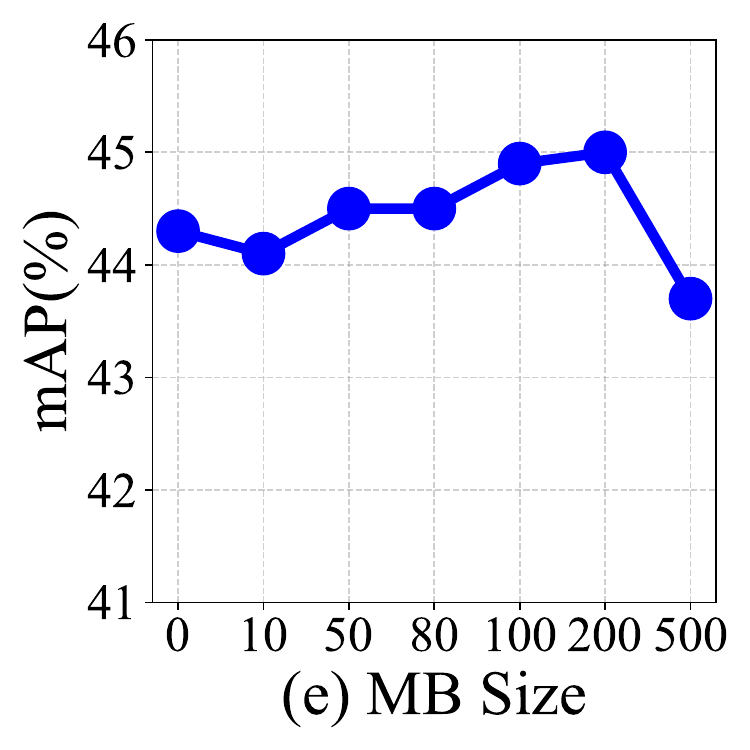}
    \includegraphics[width=0.15\textwidth]{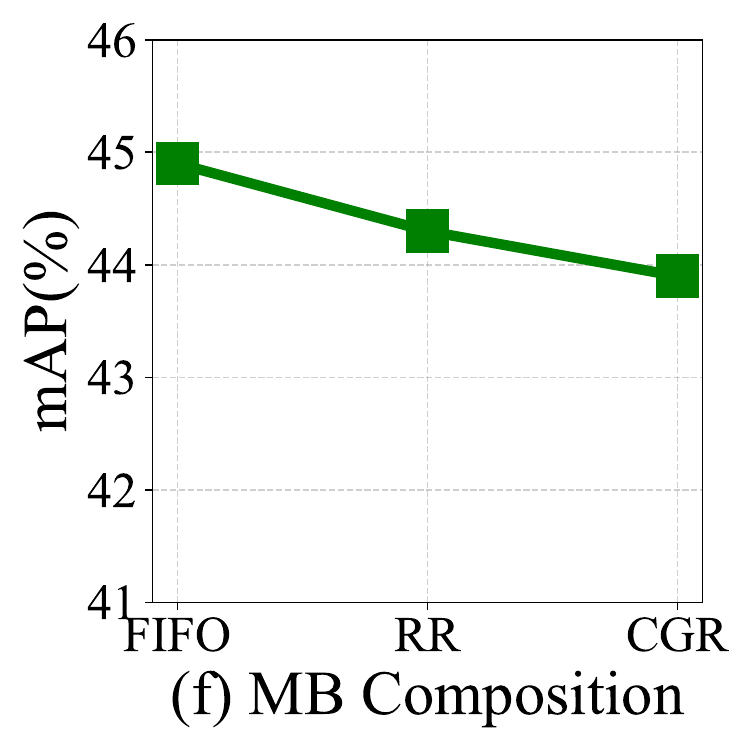}
    \caption{Experimental results from the ablation studies: (a) Influence of different features on object estimation. (b) Influence of various encoder feature fusion layers. (c) Comparisons between thresholding and matching in CMMB. (d) Influence of controlled noise levels in CMMB. (e) Influence of memory bank (MB) size. (f) Influence of memory bank (MB) composition, including first-in-first-out (FIFO), random replacement (RR), and center-guided replacement (CGR).} \label{fig:ablation_4}
\end{figure*}

\begin{figure*}[thb] \centering
    \includegraphics[width=0.15\textwidth]{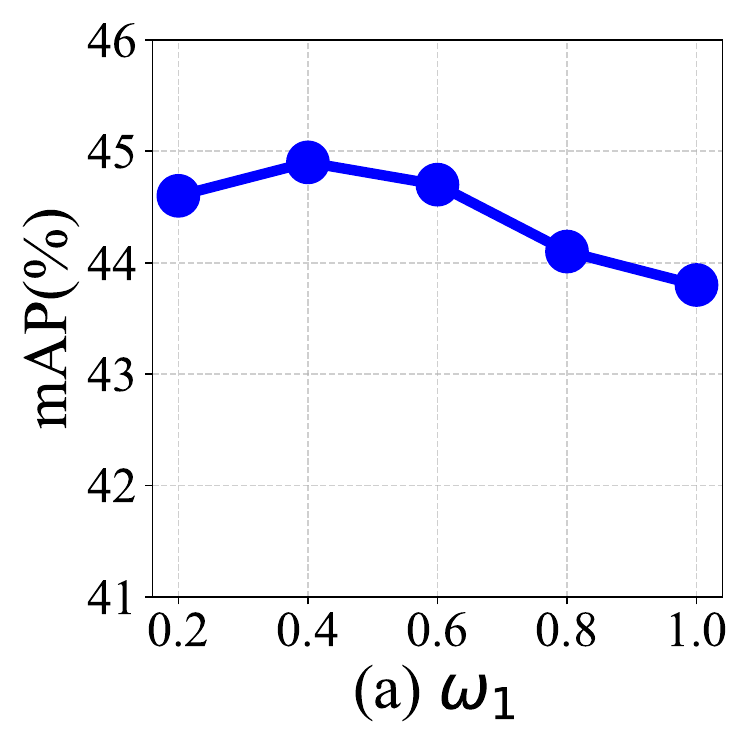}
    \includegraphics[width=0.15\textwidth]{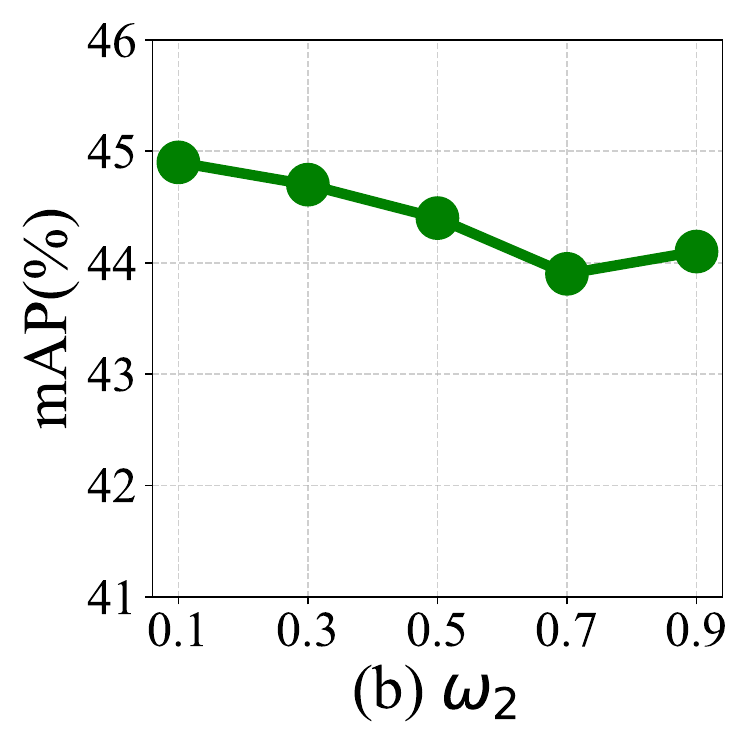}
    \includegraphics[width=0.15\textwidth]{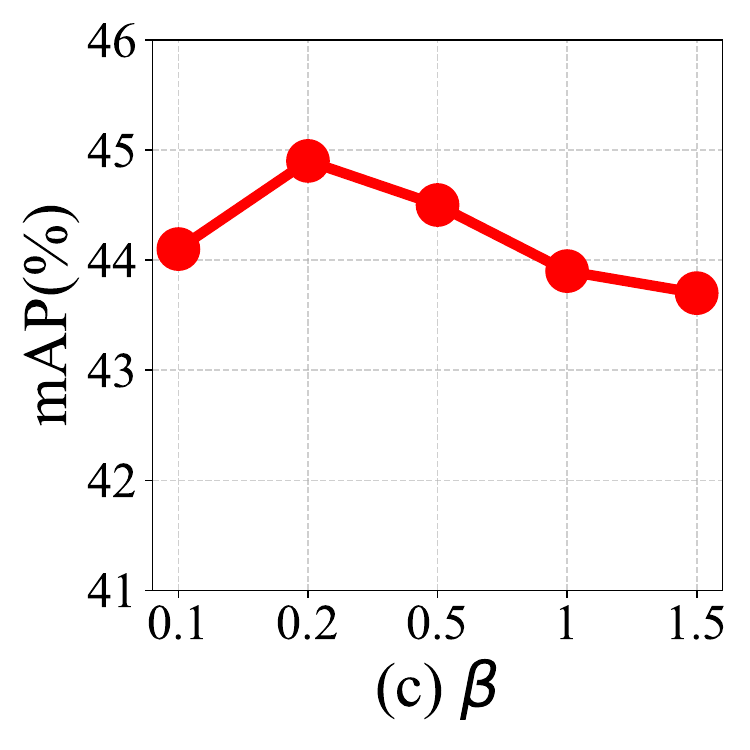}
    \includegraphics[width=0.15\textwidth]{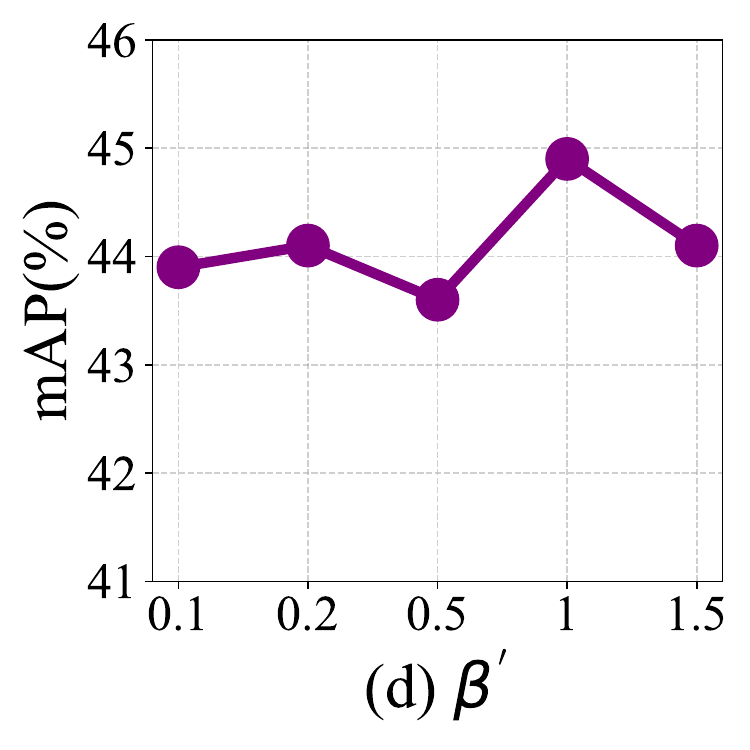}
    \includegraphics[width=0.15\textwidth]{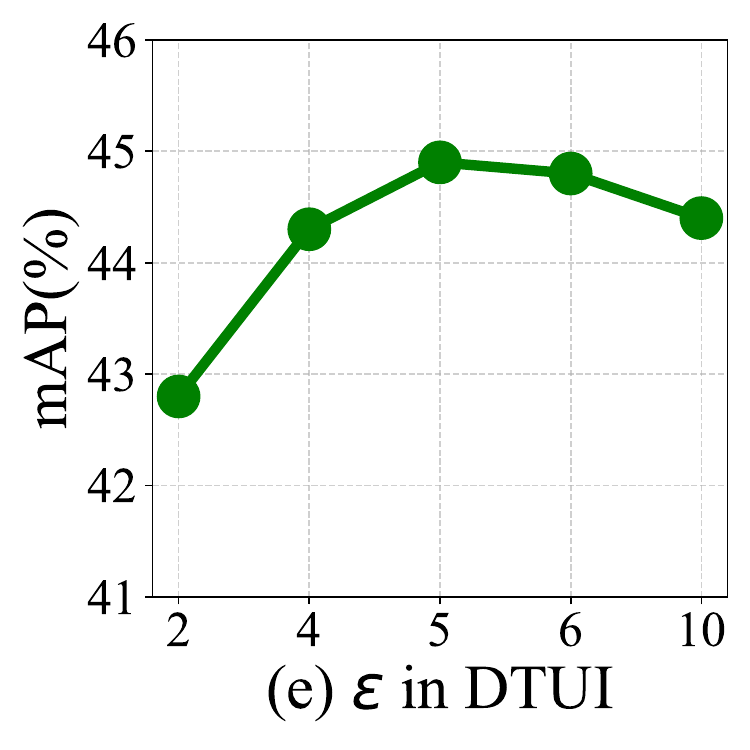}
    \includegraphics[width=0.15\textwidth]{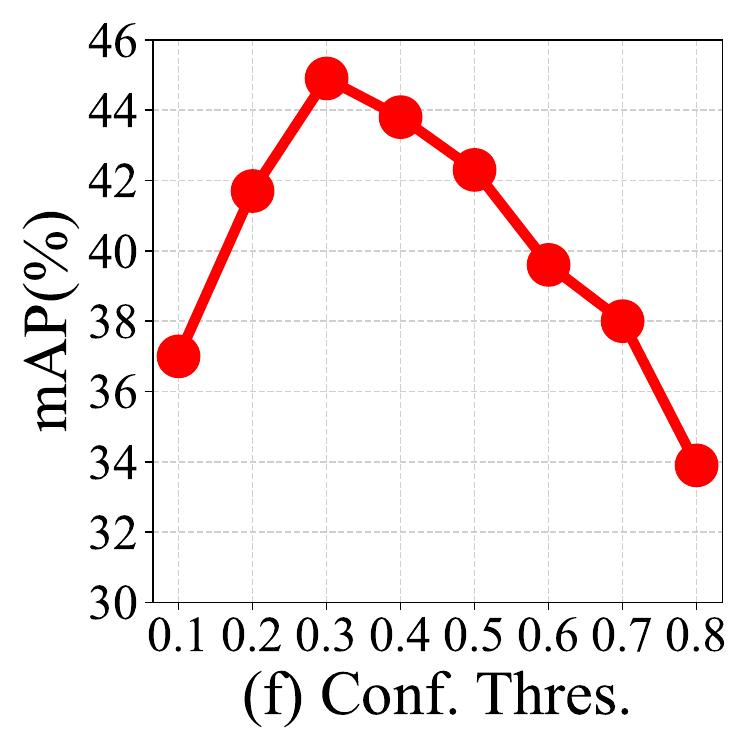}
    \caption{Hyperparameter Sensitivity Analysis. We illustrate the sensitivity of four key hyperparameters: $\omega_1$, $\omega_2$, $\beta$, $\beta’$, $\epsilon$, and $c_{\mathrm{thresh}}$ (Confidence Threshold). Each plot shows performance variation when adjusting a single hyperparameter while keeping the others fixed. The definitions of these hyperparameters are provided in Table~\ref{tab:hyperparam}.} \label{fig:ablation_parameter_sensitivity}
\end{figure*}

\noindent \textbf{Effectiveness Across DETR Variants.}
Since our main experiments are based on Deformable DETR~\cite{zhu2020deformable}, we further investigate the adaptability of our method by evaluating it on different DETR variants, including DINO DETR~\cite{zhang2022dino} and RT DETR~\cite{zhao2024detrs}. Notably, both DINO DETR and RT DETR incorporate a query selection mechanism that relies on ground truth labels, which contradicts the SFOD setting. To address this limitation, we disable query selection during teacher forward propagation, subsequently leveraging pseudo-labels for student query selection and SFOD training. Experimental results in Tab.~\ref{tab:variants} demonstrate that our method consistently surpasses baseline by a large margin across all three DETR variants, highlighting both its effectiveness and adaptability.

\noindent \textbf{Hyperparameter Sensitivity.} To evaluate the robustness of the proposed components, we conduct a hyperparameter sensitivity analysis on several key hyperparameters: (1) $\omega_1$ and $\omega_2$, which balance the extra losses, (2) $\beta$ and $\beta’$, which control the weights for query-based sample reweighting and feature distillation, respectively. As shown in Fig.~\ref{fig:ablation_parameter_sensitivity}, the proposed method demonstrates strong robustness to small variations in these hyperparameters, further validating its effectiveness.

\noindent \textbf{Contrastive Pair Construction Strategy Performance and Noise Robustness.} We evaluate two contrastive pair construction strategies in CMMB: Threshold and Matching. For Threshold, we first associate student predictions with teacher pseudo‑labels by IoU‑based assignment, then filter the matched pairs using a student confidence threshold. For Matching, we instead perform a global one‑to‑one bipartite assignment between student and teacher outputs. As shown in Fig.~\ref{fig:ablation_4}(c), the threshold strategy yields mAPs of 43.7/44.4/44.3 at thresholds 0.3/0.6/0.8, while matching achieves 44.9. This indicates that relying on student confidence introduces bias from unstable predictions, whereas bipartite matching with teacher guidance leads to more reliable contrastive pairs. We further test robustness by injecting label noise, as shown in Fig.~\ref{fig:ablation_4}(d). At 10\% noise, both remain close (44.7 vs. 44.5), but as noise rises to 70\%, threshold drops to 42.1 while matching stays at 43.3. This demonstrates that Hungarian matching’s global assignment mitigates mislabeled samples and maintains a purer memory bank, whereas local thresholding admits more noisy entries.

\noindent  \textbf{Dynamic MT Updating.} Several recent studies explore dynamic teacher updates based on uncertainty estimation, leveraging techniques such as logits variance (\textbf{Var})\cite{khanh2024dynamic}, prediction entropy (\textbf{Ent})\cite{zhao2023towards}, and Soft Neighborhood Density (\textbf{SND})~\cite{zhao2023towards,saito2021tune}. To further investigate this, we conduct an ablation study comparing three different updating strategies:
\begin{enumerate}
    \item \textbf{Fixed intervals}. Teacher updates after constant iterations, with the meta interval defining the update period.
    \item \textbf{Uncertainty-based updates (Var, Ent, SND)}. Updates occur when the uncertainty metric decreases, with the meta interval as the max interval, following DRU~\cite{khanh2024dynamic}.
    \item \textbf{Proposed DTUI}. The proposed updating mechanism where the meta interval corresponds to $\delta$ in Eq.~(~\ref{equ:iema}).
\end{enumerate}


Following prior work~\cite{khanh2024dynamic}, we set the meta interval to 5 for uncertainty-based updates (Var, Ent, SND). As shown in Tab.~\ref{tab:abl_teacherupdate}, both DTUI and uncertainty-based strategies outperform the vanilla Mean Teacher, with DTUI yielding the best results. However, uncertainty-based methods show limited gains unless combined with techniques like DRU’s student retraining\cite{khanh2024dynamic}, possibly due to noisy background queries and unstable proposals in DETR. We also observe that overly large or small intervals offer minimal benefits, highlighting the importance of a balanced update frequency in DTUI. Additionally, Fig.~\ref{fig:ablation_parameter_sensitivity}(e) shows that DTUI is robust to small changes in $\epsilon$, which controls interval growth. Performance only degrades when $\epsilon$ is too large (e.g., 2), delaying necessary teacher updates.

\noindent \textbf{Objectness Estimation Strategy.} To assess the superiority of OSSR under different objectness estimation strategies, we conduct an ablation study comparing the following approaches:
\begin{enumerate}
    \item \textbf{Backbone}: Utilizing backbone features alone.
    \item \textbf{E}: Utilizing encoder features alone.
    \item \textbf{E+AQ}: Utilizing encoder features fused with assigned queries (\textit{i.e.}, queries matched with pseudo-labels via bipartite matching).
    \item \textbf{E+Q}: Utilizing encoder features fused with all queries.
\end{enumerate}
As shown in Fig.~\ref{fig:ablation_4}(a), the highest performance is achieved when using encoder features fused with assigned queries (E+AQ). This result highlights the effectiveness of objectness estimation and query-fused feature weighting. Additionally, these quantitative findings complement the visualization results in Fig.~\ref{fig:vis_attention}, further validating the impact of our approach.

\noindent \textbf{Pseudo-labeling Threshold.} We also conduct an ablation study on the pseudo-labeling threshold $c_{\mathrm{thresh}}$, which is set to 0.3 in our main experiments. As shown in Fig.~\ref{fig:ablation_parameter_sensitivity}(f), both excessively low and high thresholds lead to suboptimal detection performance, while a threshold around 0.3 yields the best results. This finding is consistent with DRU~\cite{khanh2024dynamic}, which also uses a 0.3 threshold for DETR-based pseudo-labeling.

\noindent \textbf{Memory Bank in CMMB.}
We conduct ablations on both memory bank size and update strategy. Sizes range from 0 (disabling the memory bank) to several capacities per class, and update strategies include FIFO, random replacement (RR), and center‑guided replacement (CGR), where a new feature replaces the entry farthest from the current class center. As shown in Fig.~\ref{fig:ablation_4}(e) and (f), moderate sizes (around 100–200 entries) give the best mAP, while very small or very large sizes slightly reduce performance. Among update strategies, FIFO performs best, with RR and CGR yielding slightly lower results. These findings suggest that a balanced memory size and stable FIFO updates provide the most effective composition for contrastive learning in CMMB.


\begin{table}[!t]
\centering\footnotesize
\caption{Ablation study on component analysis. The reported \textit{Time} denotes the per-iteration processing time, and \textit{$\Delta$Time} indicates the relative increase compared with the MT+DTUI baseline.}
\resizebox{0.95\linewidth}{!}{%
\begin{tabular}{cccc|ccc}
\toprule[1.5pt]
MT+DTUI & CMMB & OSSR & UQFD & mAP & \textit{Time/ms} & \textit{$\Delta$Time/\%} \\
\midrule
\usym{1F5F8} &   &   &   & 41.7 & 387.5 & --   \\
\usym{1F5F8} &   &   & \usym{1F5F8} & 42.1 & 508.7 & 23.8 \\
\usym{1F5F8} &   & \usym{1F5F8} &   & 42.3 & 428.8 & 9.6  \\
\usym{1F5F8} & \usym{1F5F8} &   &   & 43.5 & 416.2 & 6.9  \\
\usym{1F5F8} &   & \usym{1F5F8} & \usym{1F5F8} & 43.2 & 517.7 & 25.1 \\
\usym{1F5F8} & \usym{1F5F8} &   & \usym{1F5F8} & 44.2 & 519.9 & 25.5 \\
\usym{1F5F8} & \usym{1F5F8} & \usym{1F5F8} &   & 44.4 & 477.2 & 18.8 \\
\usym{1F5F8} & \usym{1F5F8} & \usym{1F5F8} & \usym{1F5F8} & \textbf{44.9} & 542.1 & 28.5 \\
\bottomrule[1.5pt]
\end{tabular}%
}
\label{tab:abl_components}
\end{table}



\noindent \textbf{Multi-scale Encoder Feature Fusion.} To assess the impact of multi-scale encoder feature fusion in OSSR, we conduct an ablation study on different fusion strategies using layers ${3,4,5}$. We experiment with fusing features from individual layers as well as multiple layers. As shown in Fig.~\ref{fig:ablation_4}(c), (1) reweighting based on lower-layer encoder features leads to suboptimal performance, as they emphasize high-level semantics over object-level attention. While fusing layer 5 with others yields only marginal gains, it confirms that multi-layer fusion provides more comprehensive features than a single layer, ultimately enhancing overall performance.

\noindent \textbf{Effectiveness over Different Detectors.}
To further illustrate the DETR-specific design, we directly transfer our method to Faster R‑CNN~\cite{ren2015faster} by replacing query and encoder features with RoI and backbone features. As shown in Tab.~\ref{tab:abl_frcnn}, the modules yield only marginal or even degraded gains compared to the clear improvements on DETR. This stems from our DETR‑specific design: OSSR and UQFD rely on query–encoder fused objectness, while RoI features lack the global semantic context of DETR queries, weakening supervision and distillation; CMMB also underperforms as it is built around learnable queries rather than local RoIs. Moreover, Hungarian matching on thousands of RPN proposals incurs extra cost and instability. These findings underscore the query‑centric nature of our approach and suggest that a promising direction for future work is to generalize the key ideas and designs of DETR‑based SFOD so that they can be effectively adapted to a broader range of detector architectures.

\begin{table}[!t]
\centering\footnotesize
\caption{Ablation Study on Dynamic MT(Mean Teacher) updating analysis. ``Var'', ``Ent'', and ``SND'' denote logits variance, prediction entropy, and Soft Neighborhood Density, respectively, as stated in Sec.~\ref{ssec:abl}. \textbf{MI} represents Meta Interval. }
\resizebox{1.0\linewidth}{!}{%
\begin{tabular}{c|cccc|ccc|ccc}
\toprule[1.5pt]
Strategy & \multicolumn{4}{c|}{Fixed intervals} & Var & Ent & SND & \multicolumn{3}{c}{DTUI} \\
\midrule
MI & 1 & 2 & 5 & 10 & 5 & 5 & 5 & 2 & 5 & 10 \\
\midrule
mAP & 41.4 & 42.4 & 44.1 & 41.7 & 44.4 & 43.6 & 43.8 & 43.5 & \textbf{44.9} & 43.1 \\
\bottomrule[1.5pt]
\end{tabular}%
}
\label{tab:abl_teacherupdate}
\end{table}



\begin{table}[!t]
\centering\footnotesize
\caption{Ablation Study on different module combinations over different detectors.}
\resizebox{0.85\linewidth}{!}{%
 \begin{tabular}{c|cccc|c}
    \toprule[1.5pt]
    Detector & MT+DTUI & CMMB & OSSR & UQFD & mAP \\
    \midrule
    \multirow{5}{*}{FRCNN} & \usym{1F5F8} &  &  &  & 37.4 \\
     & \usym{1F5F8} & \usym{1F5F8} &  &  & 37.9 \\
     & \usym{1F5F8} &  & \usym{1F5F8} &  & 36.4 \\
     & \usym{1F5F8} &  &  & \usym{1F5F8} & 37.1 \\
     & \usym{1F5F8} & \usym{1F5F8} & \usym{1F5F8} & \usym{1F5F8} & 37.2 \\
    \midrule
    DETR & \usym{1F5F8} & \usym{1F5F8} & \usym{1F5F8} & \usym{1F5F8} & \textbf{44.9} \\
    \bottomrule[1.5pt]
 \end{tabular}%
}
\label{tab:abl_frcnn}
\end{table}

\subsection{Visualization Study} \label{ssec:vis}
To further illustrate the effectiveness of our method, we present a visualization study in this section.

\begin{figure*}[t]
\centering
\includegraphics[width=1.0\linewidth]{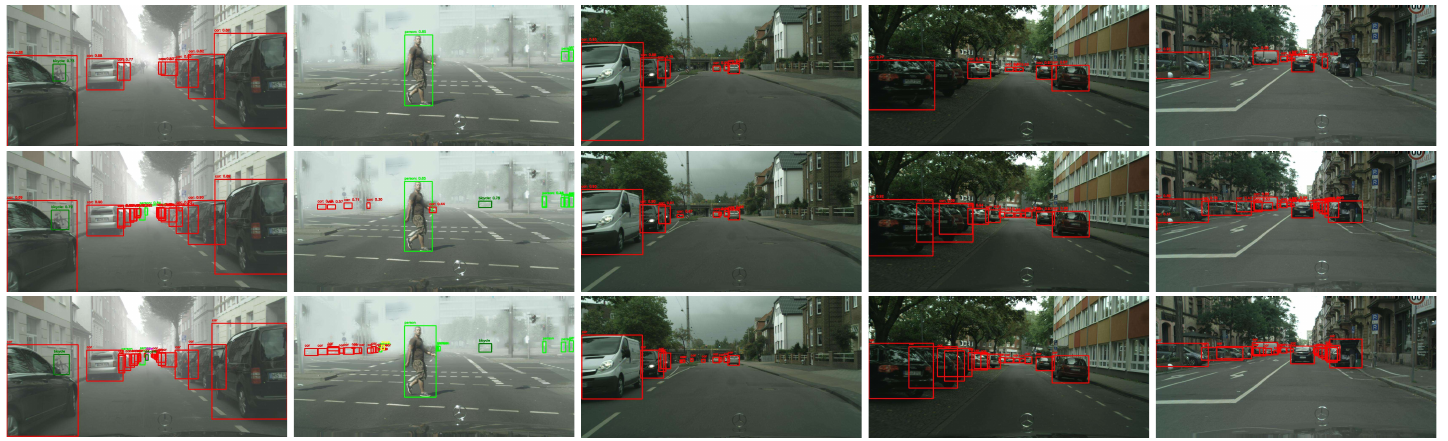}
   \caption{Visualization of detection results before and after adaptation and gt labels. From top to bottom: Source-only results, FRANCK results, and ground truth. We show both Cross-weather adaptation results and Synthetic-to-real adaptation results in this experiment.
  }\label{fig:vis_detection}
\end{figure*}

\noindent \textbf{Detection Result Visualization.} To showcase effective domain adaptation, we visualize detection results from three settings: (1) \textbf{Source-only}, (2) \textbf{FRANCK}, and (3) \textbf{Ground Truth}. As shown in Fig.~\ref{fig:vis_detection}, our method significantly improves both object localization and classification in the target domain, reducing false positives and enhancing detection quality.

\noindent \textbf{Object Attention Visualization.} To illustrate the impact of query-fused objectness scores on encoder features, we compare OSSR objectness attention generation methods, including ``\textbf{E}'', ``\textbf{E+AQ}'', and ``\textbf{E+Q}'', as outlined previously. To better visualize the weight distribution, we also subtract the scaled attention score of ``\textbf{E+Q}'' from 1, since Eq.~(~\ref{equ:wi}) assigns \textbf{high} weights to \textbf{low} attention scores.

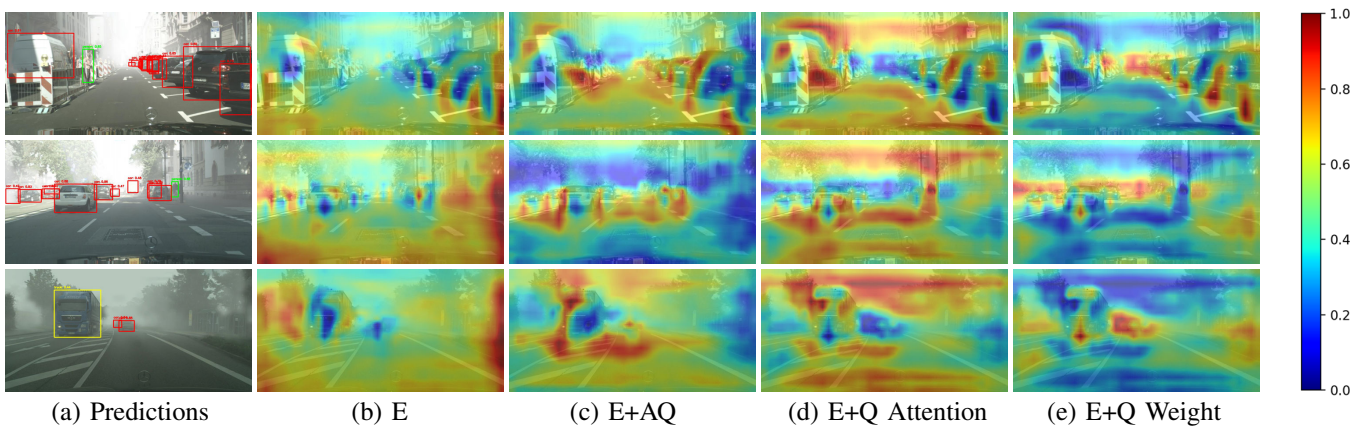
\begin{figure*}[t]
\centering
   \begin{overpic}[width=1.0\textwidth]{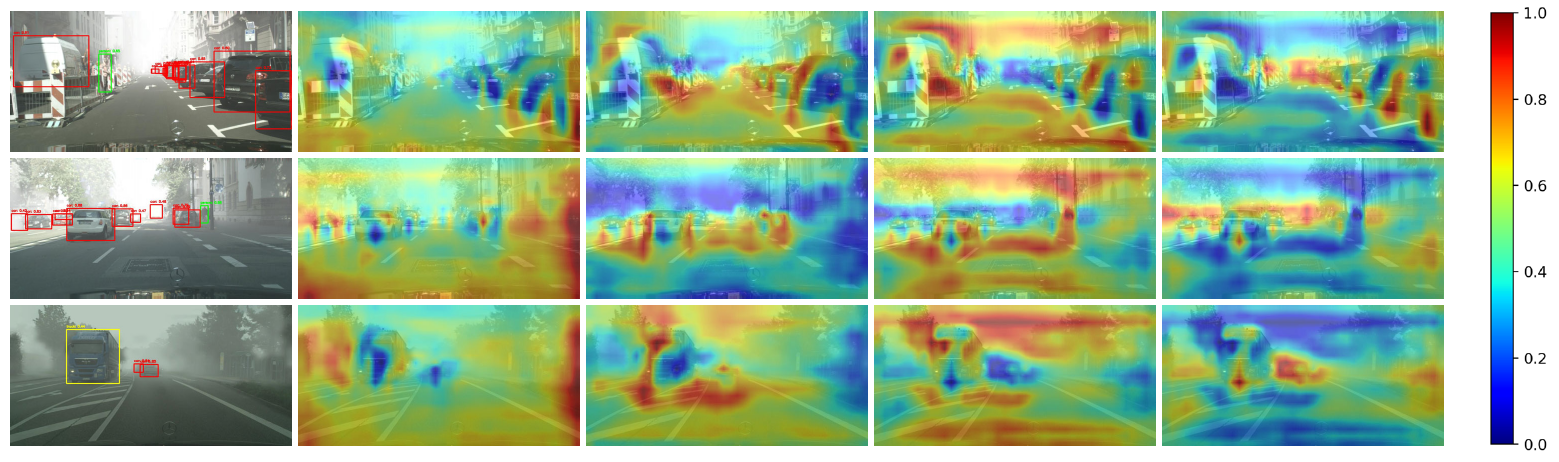} 
   \put(4,-1){(a) Predictions}
   \put(26,-1){(b) E}
   \put(42,-1){(c) E+AQ}
   \put(58,-1){(d) E+Q Attention}
   \put(77,-1){(e) E+Q Weight}
   \end{overpic}
   \caption{Visualization of object attention visualization with different strategies. ``E'' is for Encoder feature only, ``E+AQ'' is for fusion of Encoder feature and assigned query features, and ``E+Q'' is for fusion of Encoder feature and all query features. In the proposed method, we assign \textbf{high} weights to \textbf{low} E+Q attention scores and vice versa.
  }\label{fig:vis_attention}
\end{figure*}

As shown in Fig.~\ref{fig:vis_attention}, domain shifts often lead encoder features to over-focus on background regions. Query fusion mitigates this by redirecting attention toward objects, with ``E+AQ'' showing the most improvement. However, treating all queries equally (``E+Q'') can suppress attention to foreground and hard-to-detect objects, consistent with findings in DETRDistill\cite{chang2023detrdistill}. Our reweighting strategy corrects this by assigning higher weights to under-recognized areas, addressing both foreground-background and easy-hard sample imbalances. This improves feature discriminability, as further validated in Fig.~\ref{fig:ablation_4}(a).

\subsection{Limitation}
While our method achieves state-of-the-art performance, it has limitations that point to future directions. First, although a lower confidence threshold works well across varied scenarios, it may underperform in precision-critical cases. A dynamic threshold, starting low to mine potential objects and increasing later to enhance precision, could better balance recall and precision. Second, with the rise of vision foundation models (VFMs) like CLIP~\cite{radford2021learning}, incorporating VFM-guided cues (e.g., text-image similarity for refining memory bank samples) could further enhance contrastive learning, especially for applications less constrained by computation.

\section{Conclusion} \label{sec:conclusion}
In this paper, we tackle the challenge of source-free domain adaptive object detection (SFOD), specifically focusing on adapting source-pre-trained DETR networks to target domains without access to source data. To this end, we propose FRANCK, a novel framework that fully exploits DETR-specific features by incorporating four key components: (1) an Objectness Score-based Sample Reweighting (OSSR) module, (2) a Contrastive Learning with Matching-based Memory Bank (CMMB) module, (3) an Uncertainty-weighted Query-fused Feature Distillation (UQFD) module, and (4) an enhanced self-training pipeline with Dynamic Teacher Updating Interval (DTUI). Our method achieves state-of-the-art performance, surpassing previous SFOD approaches across multiple widely used benchmarks. For future work, we aim to extend our framework to more realistic scenarios, such as multi-source adaptation. We hope this work provides valuable insights and inspiration for advancing DAOD and SFOD, further contributing to the broader research community. 


\bibliographystyle{IEEEtran}
\bibliography{main} 












\end{document}